\documentclass[british,english]{IEEEtran}
\usepackage[T1]{fontenc}
\usepackage[latin9]{inputenc}
\usepackage{color}
\usepackage{babel}
\usepackage{array}
\usepackage{multirow}
\usepackage{amsmath}
\usepackage{amsthm}
\usepackage{amssymb}
\usepackage{graphicx}
\usepackage[unicode=true]
 {hyperref}

\makeatletter

\providecommand{\tabularnewline}{\\}

\newcommand{\lyxaddress}[1]{
\par {\raggedright #1
\vspace{1.4em}
\noindent\par}
}
\theoremstyle{plain}
\newtheorem{thm}{\protect\theoremname}

\@ifundefined{showcaptionsetup}{}{%
 \PassOptionsToPackage{caption=false}{subfig}}
\usepackage{subfig}
\makeatother

\addto\captionsbritish{\renewcommand{\theoremname}{Theorem}}
\addto\captionsenglish{\renewcommand{\theoremname}{Theorem}}
\providecommand{\theoremname}{Theorem}

\begin{document}

\title{Neural Component Analysis for Fault Detection}

\author{Haitao Zhao}
\maketitle

\lyxaddress{\textcolor{black}{Key Laboratory of Advanced Control and Optimization
for Chemical Processes of Ministry of Education}, School of Information
Science and Engineering, East China University of Science and Technology}
\begin{abstract}
Principal component analysis (PCA) is largely adopted for chemical
process monitoring and numerous PCA-based systems have been developed
to solve various fault detection and diagnosis problems. Since PCA-based
methods assume that the monitored process is linear, nonlinear PCA
models, such as autoencoder models and kernel principal component
analysis (KPCA), has been proposed and applied to nonlinear process
monitoring. However, KPCA-based methods need to perform eigen-decomposition
(ED) on the kernel Gram matrix whose dimensions depend on the number
of training data. Moreover, prefixed kernel parameters cannot be most
effective for different faults which may need different parameters
to maximize their respective detection performances. Autoencoder models
lack the consideration of orthogonal constraints which is crucial
for PCA-based algorithms. To address these problems, this paper proposes
a novel nonlinear method, called neural component analysis (NCA),
which intends to train a feedforward neural work with orthogonal constraints
such as those used in PCA. NCA can adaptively learn its parameters
through backpropagation and the dimensionality of the nonlinear features
has no relationship with the number of training samples. \textcolor{black}{Extensive
experimental results on the Tennessee Eastman (TE) benchmark process
show the superiority of NCA in terms of missed detection rate (MDR)
and false alarm rate (FAR)}. The source code of NCA can be found in
\href{https://github.com/haitaozhao/Neural-Component-Analysis.git}{https://github.com/haitaozhao/Neural-Component-Analysis.git}.

Note to Practitioner: Online monitoring for chemical process has been
considered as a critical and hard task in real industrial applications.
In this paper, an innovative method called neural component analysis
(NCA) is proposed for fault detection. NCA is a unified model including
a nonlinear encoder and a linear decoder. Due to its simple and intuitive
format, NCA has superior performance in both computational efficiency
and fault detection which makes it suitable for process monitoring
in real industrial applications. Moreover, the experimental results
presented can be reproduced effortlessly.
\end{abstract}


\section{Introduction}

\selectlanguage{british}%
Monitoring process conditions is crucial to its normal operation \cite{qin2012survey}.
Over last decades, data-driven multivariate statistical process monitoring
(MSPM) has been widely applied to fault diagnosis for industrial process
operations and production results \cite{macgregor2012monitoring,yin2012comparison}.
Due to the data-based nature of MSPM, it is relatively convenient
to apply to real processes of large scale comparing to other methods
based on theoretical modelling or rigorous derivation of process systems
\cite{GeSong-5,FeitalKruger-11}. 

The task of MSPM is challenging mainly due to the ``curse of dimensionality''
problem and the ``data rich but information poor'' problem. Many
methods have been proposed to transform original high dimensional
process data into a lower dimensional feature space and then performing
fault detection or fault diagnosis in that feature space \cite{askarian2016fault}.
Principal component analysis (PCA) \cite{gao2016improved,7956215,7460929}
is one of the most widely used linear techniques for fault detection.
Due to orthogonal linear projection, PCA separates data information
into two subspaces: a significant subspace which contains most variation
in training data and a residual subspace which includes noises or
outliers in training data. 

PCA-based methods to inherently nonlinear processes may lead to unreliable
and inefficient fault detection, since a linear transformation is
hard to tackle the nonlinear relationship between different process
variables \cite{luo201611,7310889}. To deal with this problem, various
nonlinear extensions of PCA have been proposed for fault detection.
These extensions can be divided into 3 categories. 

The first category is kernel approaches. Kernel PCA (KPCA) is one
of the mostly used kernel approaches for fault detection \cite{kpcaMANSOURI2016334}.
KPCA implicitly maps the data from an input space into some high dimensional
nonlinear feature sapce, where linear PCA can be applied. KPCA need
to perform eigen decomposition (ED) on the kernel Gram matrix whose
size is the square of the number of data points. When there are too
many data points, the calculation of ED becomes hard to perform \cite{robust_self_supervised}.
Moreover, KPCA need to determine the kernel and the associated parameters
in advance.

The second category is based on linear approximation of nonlinear
process. In the linear approximation, several local linear models
are constructed and then integrated by Bayesian inference \cite{GE2010676}.
Linear approximation is simple and easy to realize, but it may not
be able to handle strong nonlinearities in the process. 

The third category is neural-network-based models, such as robust
autoencoder (RAE) \cite{robust_self_supervised} and autoassociative
neural network \cite{AssociatedNN}. These models train a feedforward
neural network to perform the identity encoding, where the inputs
and the outputs of the network are the same. The network contains
an internal ``bottleneck'' layer (containing fewer nodes than the
output and input layers) for feature extraction. In autoassociative
neural network \cite{AssociatedNN}, the mapping from the input layer
to the ``bottleneck'' layer can be considered as encoding, while
de-mapping from the ``bottleneck'' layer to the output layer can
be considered as decoding. Although the encoding can deal with the
nonlinearities present in the data, it has no consideration of the
orthogonal constraints used in PCA. 

\begin{figure}
\subfloat[Illustration of Vector $X$ on two orthogonal directions.\label{fig:sub_orth}]{\includegraphics[scale=0.43]{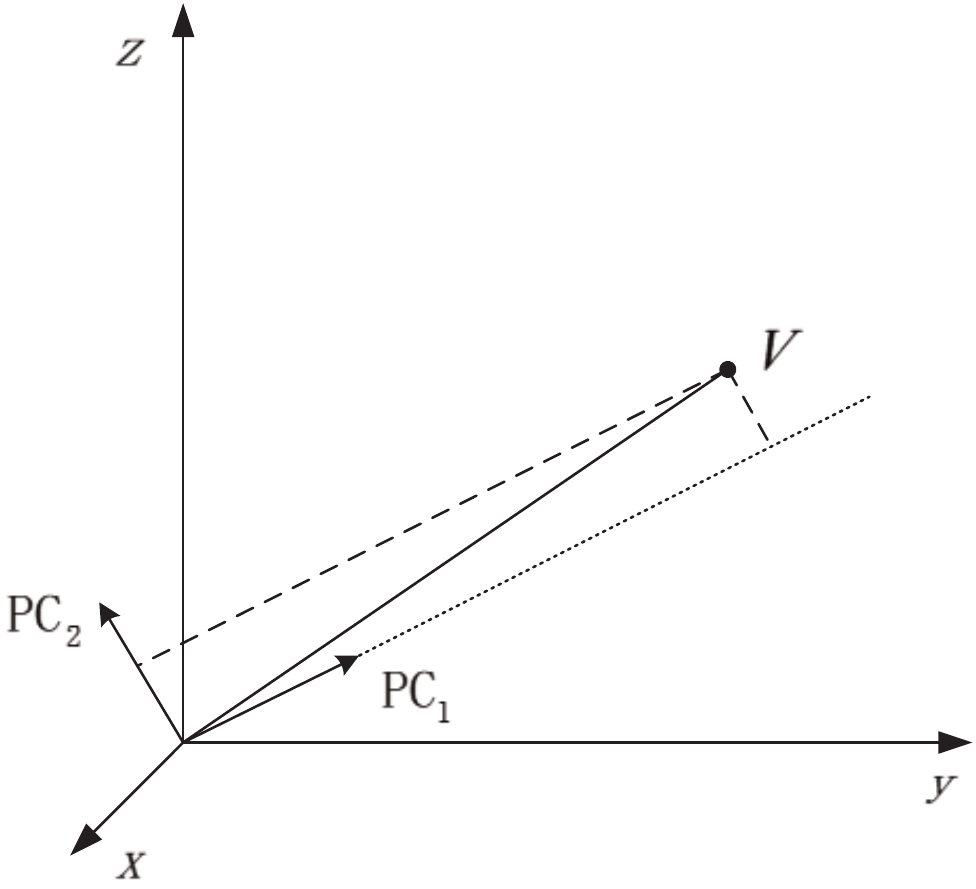}}\hspace{0.5cm}\subfloat[Illustration of Vector $X$ on two non-orthogonal directions.\label{fig:sub_nonorth}]{\includegraphics[scale=0.43]{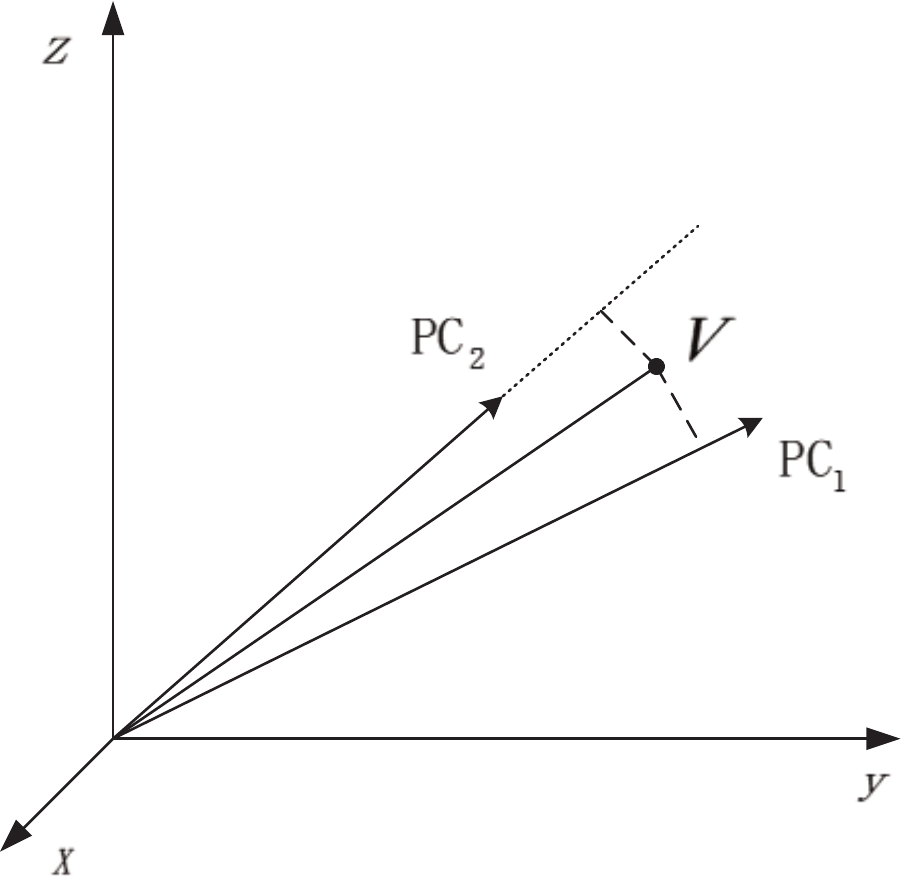}}\caption{Projections of Vector $X$ on non-orthogonal and orthogonal directions
respectively.\label{fig:orth-nonorth}}
\end{figure}

Recent years, a new trend in neural-network-based techniques known
as deep learning has become popular in artificial intelligence and
machine learning \cite{Hinton504}. Deep-learning-based models are
widely used in unsupervised training to learn the representation of
original data. Although these models are often derived and viewed
as extensions of PCA, all of them lack the consideration of the orthogonal
constraints used in PCA. The orthogonal constraints are quite important,
since they can largely reduce the correlations between extracted features.
Figure \ref{fig:orth-nonorth} shows simple plots of features of Vector
$V$ obtained by orthogonal projections and non-orthogonal projections
respectively. From Figure \ref{fig:sub_nonorth} it is easy to find
${\rm PC}_{1}$ and ${\rm PC}_{2}$ are largely correlated. It means
that the extracted features contain redundant information and may
distort the reconstruction of the original vector \cite{olpp_caideng}.

\selectlanguage{english}%

Motivated by the above analysis, this paper proposes a novel \foreignlanguage{british}{unified
model, called }neural component analysis (NCA), for fault detection.
NCA firstly utilizes a nonlinear neural network as an encoder to extract
features. Then linear orthogonal transformation is adopted to decode
the features to the original data space. Finaly, this unified model
is trained by minimizing the reconstruction error between original
data and the decoded data. After training, NCA can be used as an unsupervised
learning method to extract the key features of process data. In this
paper, Hotelling $T^{2}$ statistic and the squared prediction error
(SPE) statistic are used for fault detection. The merits of the proposed
NCA method is demonstrated by both theoretical analysis and case studies
on\textcolor{black}{{} the Tennessee Eastman (TE) benchmark process.}

\section{Autoencoder and PCA}

An autoencoder model is an artificial neural network adopted for unsupervised
feature extraction \cite{LIOU201484}. An autoencoder model tries
to learn a representation (encoding) for original data, specifically
for the purpose of dimensionality reduction. Recently, due to the
research works in deep learning, the autoencoder concept has be widely
accepted for generative models of data \cite{2013arXiv1312}.

\begin{figure}
\begin{centering}
\textcolor{black}{\includegraphics[scale=0.7]{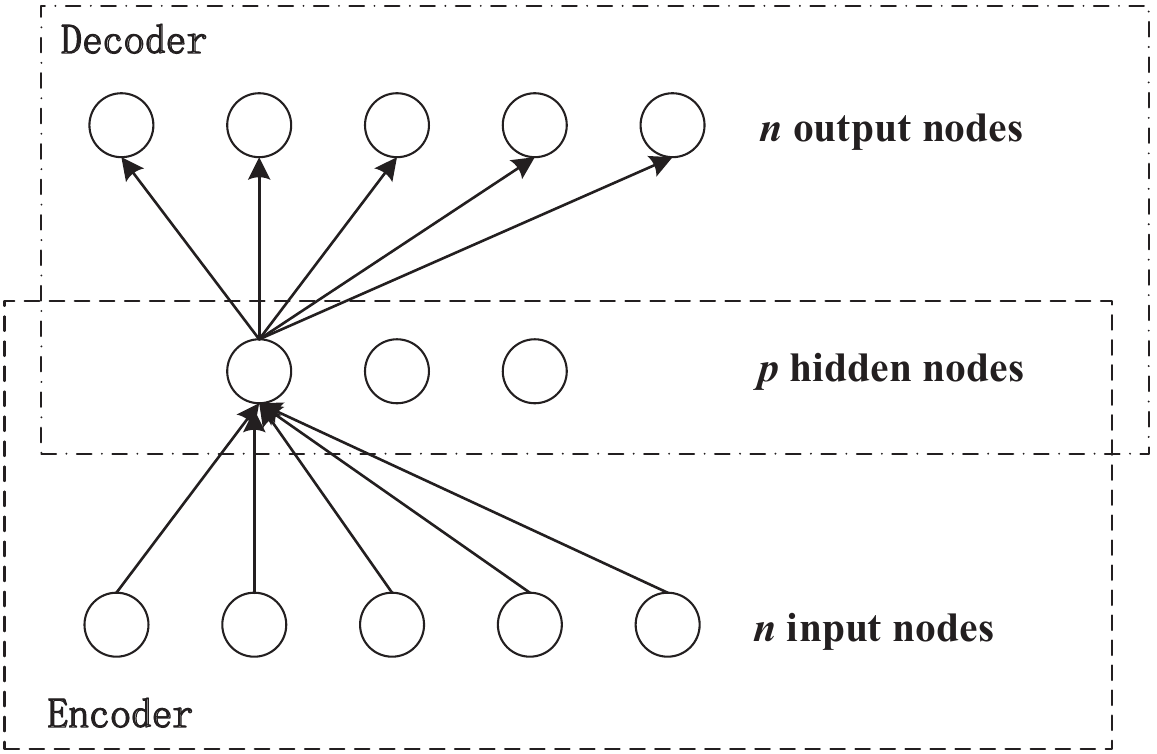}}
\par\end{centering}
\textcolor{black}{\caption{The autoencoder model. \label{fig:nn_figure}}
}
\end{figure}

We assume that there are $N$ input samples ${\bf x}_{i}=[x_{i1},x_{i2},\cdots,x_{in}]\in\mathbb{R}^{1\times n}$
$(i=1,2,\cdots,N)$. The simplest structure of an autoencoder model
is a feedforward neural network which consists of one input layer
with $n$ inputs, one hidden layer with $p$ $(p<n)$ units and one
output layer with the same number of nodes as the input layer (see
Figure \ref{fig:nn_figure}). The purpose of this structure is to
reconstruct its own inputs. Therefore, an autoencoder model belongs
to unsupervised learning.

\textcolor{black}{}

An autoencoder model includes both the encoder part and the decoder
part, which can be defined as transitions $\alpha$ and $\beta$,
such that:
\[
\begin{array}{ccc}
\alpha & \text{:} & \mathbb{R}^{1\times n}\rightarrow\mathcal{F}\\
\beta & : & \mathcal{F\rightarrow\text{\ensuremath{\mathbb{R}}}}^{1\times n}
\end{array}
\]

\begin{equation}
\alpha,\beta=\arg\min_{\alpha,\beta}\sum_{i=1}^{N}\left\Vert {\bf x}_{i}-\beta\left(\alpha\left({\bf x}_{i}\right)\right)\right\Vert ^{2}.\label{eq:autoencoder}
\end{equation}

In this paper, we only consider the case $\mathcal{F=}\mathbb{R}^{1\times p}$.
The encoder part takes the input ${\bf x}$ and maps it to ${\bf z}\in\mathbb{R}^{1\times p}$:

\[
{\bf z}=\alpha({\bf x})\triangleq\sigma\left(W{\bf x}+{\bf b}\right).
\]
This feature ${\bf z}$ is usually referred to as \textit{latent code,
latent feature} or \textit{latent representation}. Here, $\sigma(\cdot)$
is an element-wise activation function such as a sigmoid function
or a hyperbolic tangent function. Matrix $W$ is a parameter matrix
and ${\bf b}$ is a bias vector. 

The decoder part maps the feature vector ${\bf z}$ to the reconstruction
${\bf \tilde{{\bf x}}}$ of the same dimensionality as ${\bf x}$:
\[
\tilde{{\bf x}}=\beta({\bf z})\triangleq\tilde{\sigma}\left(\tilde{W}{\bf z}+\tilde{{\bf b}}\right)
\]
where $\tilde{\sigma}$, $\tilde{W}$ and $\tilde{{\bf b}}$ for the
decoder part may be different from the corresponding $\sigma$, $W$
and ${\bf b}$ for the encoder part, depending on the applications
of the autoencoder model.

In order to learn the parameters in the activation functions, an autoencoder
model is often trained to minimize reconstruction error:
\[
W,{\bf b};\tilde{W},\tilde{{\bf b}}=\arg\min_{W,{\bf b};\tilde{W},\tilde{{\bf b}}}\sum_{i=1}^{N}\left\Vert {\bf x}_{i}-\tilde{\sigma}\left(\tilde{W}\sigma\left(W{\bf x}+{\bf b}\right)+\tilde{{\bf b}}\right)\right\Vert ^{2}.
\]

If linear activation functions are used, the optimal solution to an
autoencoder is strongly related to PCA \cite{Chicco}. In this case,
Equation (\ref{eq:autoencoder}) can be written as 
\[
A,B=\arg\min_{A,B}\sum_{i=1}^{N}\left\Vert {\bf x}_{i}-{\bf x}_{i}AB^{T}\right\Vert ^{2}
\]
or
\begin{equation}
A,B=\arg\min_{A,B}\left\Vert X-XAB^{T}\right\Vert _{F}^{2}\label{eq:PCA_obj}
\end{equation}
where $X=\left[\begin{array}{c}
{\bf x}_{1}\\
{\bf x}_{2}\\
\vdots\\
{\bf x}_{N}
\end{array}\right]$ and $\left\Vert \cdot\right\Vert _{F}$ is the Frobenius norm. $A\in\mathbb{R}^{n\times p}$
and $B\in\mathbb{R}^{n\times p}$ are two linear transformation matrices.
Baldi and Hornik \cite{BALDI198953} showed that, if the covariance
matrix $\Sigma_{x}$ associated with the data $X$ is invertible,
a unique local and global minimum to Equation (\ref{eq:PCA_obj})
corresponding to an orthogonal projection onto the subspace spanned
by the first principal eigenvectors of the covariance matrix $\Sigma_{x}$.
More precisely, the optimal solutions to (\ref{eq:PCA_obj}) can be
obtain by $A=B=U_{n\times p}$, where $U_{n\times p}=[{\bf u}_{1},{\bf u}_{2},\cdots,{\bf u}_{p}]$
are the $p$ eigenvectors corresponding to the first $p$ largest
eigenvalues of the covariance matrix $\Sigma_{x}$. In PCA, ${\bf u}_{1},{\bf u}_{2},\cdots,{\bf u}_{p}$
are defined as loading vectors or principal components and $T=XU$
is the score matrix corresponding to the loading matrix $U$.

Although there is no orthogonal constraints of the transformation
matrices $A$ or $B$, the solutions to Equation (\ref{eq:PCA_obj})
are composed by orthogonormal bases. Due to the orthogonal decomposition,
PCA can transform an original data space into two orthogonal subspaces,
\foreignlanguage{british}{a principal subspace which contains most
variation in original data and a residual subspace which includes
noises or outliers}. $T^{2}$ statistic and SPE statistic are often
adopted as indicators for fault detection corresponding to the principal
subspace and the residual subspace respectively. The orthogonal decomposition
minimizes the correlations between these two subspaces and makes the
linear reconstruction of original data with the least distortion \cite{Jolliffe,Jolliffe-17}.
Because of this orthogonal property, PCA is widely used in process
monitoring and many other fault detection methods can be considered
as the extensions of PCA \cite{qin2012survey}.

\section{Neural Component Analysis}

A nonlinear autoencoder model can be trained to extract latent features.
However, due to the lacking of the orthogonal property, the significant
information of original data and the information of \foreignlanguage{british}{noises
or outliers are largely combined in this model. An autoencoder model
turns to overfit original data and learns to capture as much information
as possible rather than reducing correlations in original data and
extracting the significant information. Because of this problem, nonlinear
autoencoder models are not used as widely as PCA.}

\selectlanguage{british}%
In the section, we propose \foreignlanguage{english}{a novel }unified
model, called \foreignlanguage{english}{neural component analysis
(NCA), for fault detection. NCA firstly utilizes a nonlinear neural
network as a encoder to extract features. Then linear orthogonal transformation
is adopted to decode the features to the original data space. In this
way, NCA can be considered as a combination of nonlinear and linear
models.}

\selectlanguage{english}%
In NCA, we use linear transformation $B$ instead of nonlinear transition
$\beta$ used in Equation (\ref{eq:autoencoder}). The optimization
problem of NCA is

\begin{align}
W,{\bf b},B & =\arg\min_{W,{\bf b},B}\sum_{i=1}^{N}\left\Vert {\bf x}_{i}-g\left({\bf x}_{i};W,{\bf b}\right)B^{T}\right\Vert ^{2}\label{eq:nca_obj}\\
 & {\rm subject\ to}\ B^{T}B=I_{p\times p}\nonumber 
\end{align}
where $g\left({\bf x}_{i};W,{\bf b}\right)$ is a neural network with
$p$ outputs. Assume $B=\left[\mathfrak{b}{}_{1},\mathfrak{b}{}_{2},\cdots,{\bf \mathfrak{b}}_{p}\right]$
where $\mathfrak{b}_{i}\in\mathbb{R}^{n\times1}$, then the orthogonormal
constraint, $B^{T}B=I$, means 
\[
{\bf \mathfrak{b}}_{i}^{T}{\bf \mathfrak{b}}_{j}=\begin{cases}
1 & i=j\\
0 & {\rm otherwise}
\end{cases}\ \ \ (i,j=1,2,\cdots,p).
\]

\begin{figure}
\begin{centering}
\textcolor{black}{\includegraphics[scale=0.65]{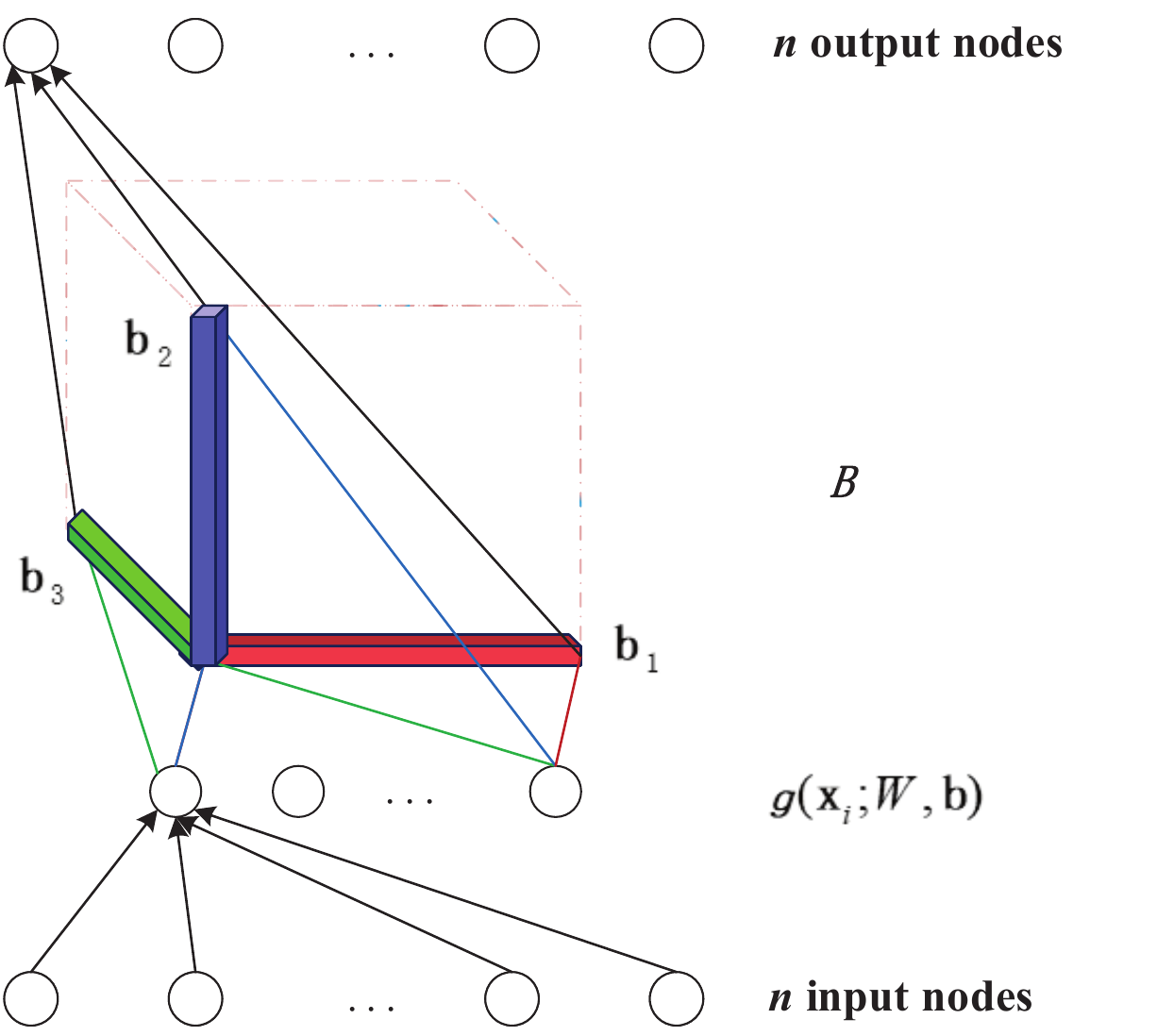}}
\par\end{centering}
\textcolor{black}{\caption{Illustration of neural component analysis (NCA). \label{fig:nca_figure}}
}
\end{figure}

Figure \ref{fig:nca_figure} illustrates the structure of NCA. Here
$B$ contains three orthogonal bases (plotted in red, blue, and green),
i.e. $B=\left[\mathfrak{b}{}_{1},\mathfrak{b}{}_{2},\mathfrak{b}_{3}\right]$.

\textcolor{black}{}

Equation (\ref{eq:nca_obj}) shows the difference between NCA and
autoencoder. Firstly, NCA is a unified model of nonlinear encoding
and linear decoding. For the linear decoding, it will be shown later
that the computation of the transformation matrix $B$ is quite simple
and no gradient-descent-based optimization is needed. Secondly, the
orthogonormal constraints $B^{T}B=I_{p\times p}$ are added to NCA.
It means that the decoding from latent features is an orthogonal reconstruction,
which can largely reduce the correlation of different variables.

Let 
\begin{equation}
G=\left[\begin{array}{c}
g\left({\bf x}_{1};W,{\bf b}\right)\\
g\left({\bf x}_{2};W,{\bf b}\right)\\
\vdots\\
g\left({\bf x}_{N};W,{\bf b}\right)
\end{array}\right]\in\mathbb{R}^{N\times p},\label{eq:G_matrix}
\end{equation}
the optimization problem of NCA in Equation (\ref{eq:nca_obj}) can
also be written as 

\begin{align}
W,{\bf b},B & =\arg\min_{W,{\bf b},B}\left\Vert X-GB^{T}\right\Vert _{F}^{2}\label{eq:nca_obj-2}\\
 & {\rm subject\ to}\ B^{T}B=I_{p\times p}.\nonumber 
\end{align}

Matrix $G$, corresponding to the score matrix of PCA, is the key
features which we want to obtain for further analysis, such as fault
detection and diagnosis. \textcolor{black}{However, it is difficult
to compute the optimal $W$, ${\bf b}$ and $B$ simultaneously since
the optimization problem in Equation (\ref{eq:nca_obj-2}) is nonconvex.
In this paper, we compute $W$, ${\bf b}$ and $B$ iteratively as
follows. We firstly fix $W$, ${\bf b}$ and obtain $G$, then $B$
can be computed by optimizing }

\textcolor{black}{
\begin{align}
B & =\arg\min_{B}\left\Vert X-GB^{T}\right\Vert _{F}^{2}\label{eq:opt_B}\\
 & {\rm subject\ to}\ B^{T}B=I_{p\times p}.\nonumber 
\end{align}
}

\textcolor{black}{Once $B$ is obtained, $W$ and ${\bf b}$ can be
updated by solving the following optimization problem:}

\textcolor{black}{
\begin{equation}
W,{\bf b}=\arg\min_{W,{\bf b}}\left\Vert X-GB^{T}\right\Vert _{F}^{2}.\label{eq:opt_w_b}
\end{equation}
}

The solution to the optimization problem in Equation (\ref{eq:opt_w_b})
can be obtained by the backpropagation algorithm \cite{5231496} which
is widely adopted in training feedforward neural networks. The solution
to Equation (\ref{eq:opt_B}) is to determine an orthogonal matrix,
that rotates $G$ to fit original data matrix $X$. In linear algebra
and statistics \cite{procruste}, Procrustes analysis is a standard
technique for geometric transformations between two matrices. The
orthogonal Procrustes problem can be viewed as a matrix approximation
problem which tries to find the optimal \textsl{rotation }or \textsl{reflection}
for the transformation of a matrix with respect to the other. Theorem
\ref{thm:Reduced-Rank-Procrustes} shows how to solve the reduced
rank Procrustes rotation problem.
\begin{thm}
\cite{TenBerge1977} \label{thm:Reduced-Rank-Procrustes} Reduced
Rank Procrustes Rotation. Let $M_{N\times n}$ and $N_{n\times p}$
be two matrices. Consider the constrained minimization problem
\begin{align*}
\hat{H} & =\arg\min_{H}\left\Vert M-NH^{T}\right\Vert ^{2}\\
 & subject\ to\ \ \ H^{T}H=I_{p\times p}.
\end{align*}
Suppose the singular value decomposition (SVD) of $M^{T}N$ is $UDV^{T}$,
then $\hat{H}=UV^{T}$.
\end{thm}
According to Theorem \ref{thm:Reduced-Rank-Procrustes}, we can design
iterative procedures of NCA to obtain $W$, ${\bf b}$ and $B$ as
follows:
\begin{enumerate}
\item Perform PCA on original data $X$ to obtain the loading matrix $U\in\mathbb{R}^{n\times p}$
and let $B=U$.
\item Fix $B$, solve the optimization problem in Equation (\ref{eq:opt_w_b})
by the backpropagation algorithm.
\item Form $G$ by Equation (\ref{eq:G_matrix}) and perform SVD on $X^{T}G$,
i.e. $X^{T}G=UDV^{T}.$
\item Compute $\hat{B}=UV^{T}$. 
\item If $\left\Vert B-\hat{B}\right\Vert <\epsilon$, break; 

else, let $B=\hat{B}$ and go to Step 2.
\item Output $W$, ${\bf b}$ and $B$.
\end{enumerate}
After training, $g\left({\bf x};W,{\bf b}\right)$ can be used for
further feature extraction. 

\section{\textcolor{black}{Fault detection based on NCA}}

\textcolor{black}{A novel fault detection method based on NCA is developed
in this section. The implementation procedures are given as follows.
Firstly in the modeling stage, the process data are collected under
normal process conditions and scaled by each variable; Then NCA is
performed to obtain the neural network }$g\left({\bf x};W,{\bf b}\right)$
\textcolor{black}{for nonlinear feature extraction.  Finally Hotelling
$T^{2}$ and the squared prediction error (SPE) statistics are used
for fault detection.  }

Let ${\bf g}_{i}=g\left({\bf x}_{i};W,{\bf b}\right)$ be the nonlinear
latent features of ${\bf x}_{i}$ $(i=1,2,\cdots,N)$ and $\Sigma_{g}$
is the covariance matrix associated with the features $G=\left[{\bf g}_{1}^{T},{\bf g}_{2}^{T},\cdots,{\bf g}_{N}^{T}\right]^{T}$.
\textcolor{black}{$T^{2}$ statistic of ${\bf g}_{i}$ is computed
as follows:
\begin{equation}
T_{i}^{2}={\bf g}_{i}\Sigma_{g}^{-1}{\bf g}_{i}^{T}.\label{eq:T2_statistic}
\end{equation}
SPE statistic of feature ${\bf g}_{i}$ can be calculated as follows:
\begin{equation}
SPE_{i}=\left\Vert {\bf x}_{i}-{\bf g}_{i}B^{T}\right\Vert ^{2}\label{eq:SPE statistic}
\end{equation}
}

Because of no prior information available about the distribution of
${\bf g}_{i}$, we compute the confidence limit for \textcolor{black}{$T^{2}$
and $SPE$ statistics approximately by kernel density estimation (KDE)
\cite{kerneldensityestimation}. Let $T_{1}^{2},T_{2}^{2},\cdots,T_{N}^{2}$
with an unknown density $\rho(\cdot)$ are the $T^{2}$ statistics
of ${\bf g}_{1},{\bf g}_{2},\cdots,{\bf g}_{N}$. The kernel density
estimator of $T^{2}$ statistic is 
\[
\hat{\rho}_{h}(T^{2})=\frac{1}{N}\sum_{i=1}^{N}K_{h}(T^{2}-T_{i}^{2})=\frac{1}{hN}\sum_{i=1}^{N}K\left(\frac{T^{2}-T_{i}^{2}}{h}\right)
\]
where $K(\cdot)$ is a non-negative function that integrates to one
and has zero mean and $h>0$ is a bandwidth parameter. In this paper,
we take the RBF kernel for density estimation, which is given by 
\[
\hat{\rho}_{h}(T^{2})=\frac{1}{\sqrt{2\pi}hN}\sum_{i=1}^{N}{\rm exp}\left(\frac{\left(T^{2}-T_{i}^{2}\right)^{2}}{2h^{2}}\right).
\]
}

\textcolor{black}{After estimating $\hat{\rho}_{h}(T^{2})$, for the
testing statistic $T_{new}^{2}$, the following condition is checked:
If $\hat{\rho}_{h}(T_{new}^{2})<\tau$ then $\mathbf{x}_{new}$ is
normal else $\mathbf{x}_{new}$ is abnormal. The threshold $\tau$
is assigned globally and could be adjusted in order to lower the percentage
of false alarm. Practically, $\tau$ is often equal to 0.01.}

\textcolor{black}{Similarly, we also can obtain 
\[
\hat{\varrho}_{h}(SPE)=\frac{1}{\sqrt{2\pi}hN}\sum_{i=1}^{N}{\rm exp}\left(\frac{\left(SPE-SPE_{i}\right)^{2}}{2h^{2}}\right)
\]
where $SPE_{1},SPE_{2},\cdots,SPE_{N}$ with an unknown density $\varrho(\cdot)$
are the $SPE$ statistics of $X_{1},X_{2},\cdots,X_{N}$. For the
testing statistic $SPE_{new}$, the following condition is checked:
If $\hat{\varrho}_{h}(SPE_{new})<\tau$ then $\mathbf{x}_{new}$ is
normal else $\mathbf{x}_{new}$ is abnormal. }

\begin{figure}
\centering{}\textcolor{black}{\includegraphics[scale=0.57]{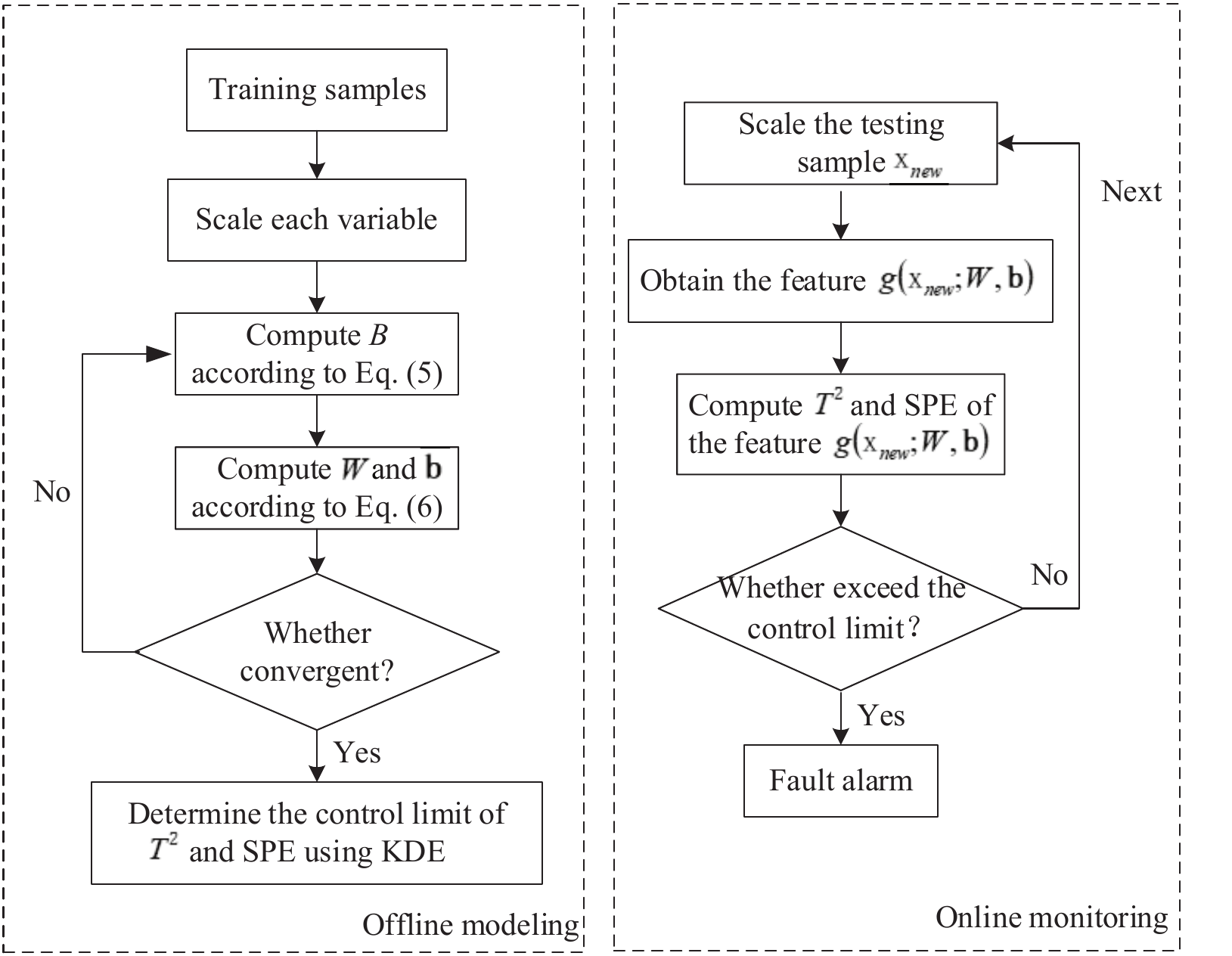}\caption{The steps of the proposed NCA method for fault detection. \label{fig:offline_online}}
}
\end{figure}

\textcolor{black}{The offline modeling and online monitoring flow
charts are shown in Figure \ref{fig:offline_online}. The procedures
of offline modeling and online monitoring are as follows:}

\textcolor{black}{}
\begin{itemize}
\item \textcolor{black}{Offline modeling:}
\end{itemize}
\begin{enumerate}
\item \textcolor{black}{Collect normal process data as the training data. }
\item \textcolor{black}{Normalize the training data by each variable with
zero mean and unit variance.}
\item \textcolor{black}{Initialize $B$ as the loading matrix which contains
the first $p$ eigenvectors of the covariance matrix of the training
data.}
\item \textcolor{black}{Fix $B$ to compute $W$ and ${\bf b}$ by training
a neural network optimizing Equation (\ref{eq:opt_w_b}).}
\item Fix \textcolor{black}{$W$ and ${\bf b}$ to compute }\textbf{\textcolor{black}{$B$
}}\textcolor{black}{by }Reduced Rank Procrustes Rotation solving Equation
(\ref{eq:opt_B}).
\item \textcolor{black}{Compute $T^{2}$ and SPE statistics of ${\bf x}_{i}$
by Equation (\ref{eq:T2_statistic}) and (\ref{eq:SPE statistic})
respectively.}
\item \textcolor{black}{Determine the control limit of $T^{2}$ and SPE
by KDE respectively.}
\end{enumerate}
\begin{itemize}
\item \textcolor{black}{Online monitoring:}
\end{itemize}
\begin{enumerate}
\item \textcolor{black}{Sample a new testing data point ${\bf x}_{new}$.
Normalize it according to the parameters of the training data.}
\item \textcolor{black}{Extract the feature }${\bf g}_{i}=g\left({\bf x}_{new};W,{\bf b}\right)$\textcolor{black}{.}
\item \textcolor{black}{Compute $T^{2}$ and SPE statistics of the feature
${\bf g}_{i}$.}
\item \textcolor{black}{Alarm if $T^{2}$ (or SPE) of the extracted feature
exceed the control limit; Otherwise, view ${\bf x}_{new}$ as a normal
data.}
\end{enumerate}
\selectlanguage{british}%

\section{Simulation and discussion }

\begin{figure}
\centering{}\includegraphics[scale=0.33]{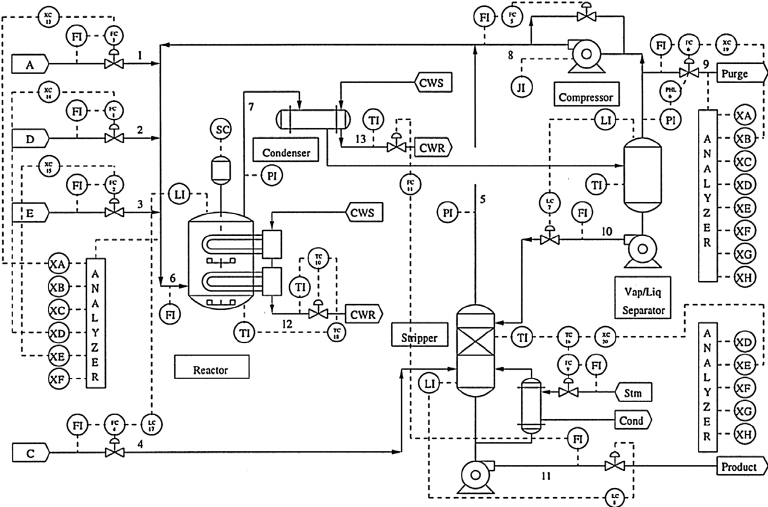}\caption{A diagram of the TEP simulator.\label{fig:A-diagram-of te}}
\end{figure}

\selectlanguage{english}%
\textcolor{black}{The Tennessee Eastman process (TEP) has been widely
used by process monitoring community as a source of publicly available
data for comparing different algorithms. The simulated TEP is mainly
based on an practical industrial process in which the kinetics, operation
and units have been altered for specific reasons. The data generated
by TEP are nonlinear, strong coupling and dynamic \cite{chiang2001fault,Lyman-42}.
There are five major units in TEP: a chemical reactor, condenser,
recycle compressor, vapor/liquid separator, and stripper. }\foreignlanguage{british}{A
flow sheet of TEP with its implemented control structure is shown
in Figure \ref{fig:A-diagram-of te}.}\textcolor{black}{{} The MATLAB
codes can be downloaded from \href{http://depts.washington.edu/control/LARRY/TE/download.html}{http://depts.washington.edu/control/LARRY/TE/download.html}.
Besides normal data, the simulator of TEP can also generate 21 different
types of faults in order to test process monitoring algorithms.}

\selectlanguage{british}%

\selectlanguage{english}%
\textcolor{black}{A total of 52 variables including 22 continuous
process measurements, 19 compositions and 11 manipulated variables}\footnote{\textcolor{black}{the agitation speed was not included because it
was not manipulated}}\textcolor{black}{{} were selected as the monitoring variables in our
experiments. The training data set contained 500 normal data. Twenty-one
different faults were generated and 960 data for each fault were chosen
for testing, in which the fault happened from 161th data to the end
of the data.}\foreignlanguage{british}{ The 21 fault modes are listed
in Table \ref{tab:TEP-fault-modes}.}

\selectlanguage{british}%
\begin{table}
\caption{TEP fault modes. \label{tab:TEP-fault-modes}}
\centering{}%
\begin{tabular}{|c|c|c|}
\hline 
Fault  & Description & Type\tabularnewline
\hline 
\hline 
1 & A/C Feed ratio, B composition constant (Stream 4) & Step\tabularnewline
\hline 
2 & B composition, A/C ratio constant (Stream 4) & Step\tabularnewline
\hline 
3 & D feed temperature (Stream 2) & Step\tabularnewline
\hline 
4 & Reactor cooling water inlet temperature & Step\tabularnewline
\hline 
5 & Condenser cooling water inlet temperature & Step\tabularnewline
\hline 
6 & A feed loss (Stream 1) & Step\tabularnewline
\hline 
7 & C header pressure loss (Stream 4) & Step\tabularnewline
\hline 
8 & A, B, C feed composition (Stream 4) & Random variation\tabularnewline
\hline 
9 & D feed temperature (Stream 2) & Random Variation\tabularnewline
\hline 
10 & C feed temperature (Stream 4) & Random Variation\tabularnewline
\hline 
11 & Reactor cooling water inlet temperature & Random Variation\tabularnewline
\hline 
12 & Condenser cooling water inlet temperature & Random Variation\tabularnewline
\hline 
13 & Reaction kinetics & Slow drift\tabularnewline
\hline 
14 & reactor cooling water valve & Sticking\tabularnewline
\hline 
15 & Condenser cooling water valve & Sticking\tabularnewline
\hline 
\selectlanguage{english}%
16\selectlanguage{british}%
 & \selectlanguage{english}%
Unknown\selectlanguage{british}%
 & \selectlanguage{english}%
Unknown\selectlanguage{british}%
\tabularnewline
\hline 
\selectlanguage{english}%
17\selectlanguage{british}%
 & \selectlanguage{english}%
Unknown\selectlanguage{british}%
 & \selectlanguage{english}%
Unknown\selectlanguage{british}%
\tabularnewline
\hline 
\selectlanguage{english}%
18\selectlanguage{british}%
 & \selectlanguage{english}%
Unknown\selectlanguage{british}%
 & \selectlanguage{english}%
Unknown\selectlanguage{british}%
\tabularnewline
\hline 
\selectlanguage{english}%
19\selectlanguage{british}%
 & \selectlanguage{english}%
Unknown\selectlanguage{british}%
 & \selectlanguage{english}%
Unknown\selectlanguage{british}%
\tabularnewline
\hline 
\selectlanguage{english}%
20\selectlanguage{british}%
 & \selectlanguage{english}%
Unknown\selectlanguage{british}%
 & \selectlanguage{english}%
Unknown\selectlanguage{british}%
\tabularnewline
\hline 
21 & Valve (Stream 4) & Constant position\tabularnewline
\hline 
\end{tabular}
\end{table}

In this paper, we compare our proposed NCA method with PCA, KPCA and
autoencoder. For KPCA, we use the most widely used Gaussian kernel
$k({\bf x}_{i},{\bf x}_{j})={\rm exp}\left(-\left\Vert {\bf x}_{i}-{\bf x}_{j}\right\Vert ^{2}/c\right)$
and select the kernel parameter $c$ as $10n\bar{\delta}$, where
$\bar{\delta}$ is the mean of standard deviations of different variables
\cite{1658299}. 

\begin{figure}
\begin{centering}
\subfloat[Visualization plot of PCA.\label{fig:sub_PCA_2d}]{\includegraphics[scale=0.25]{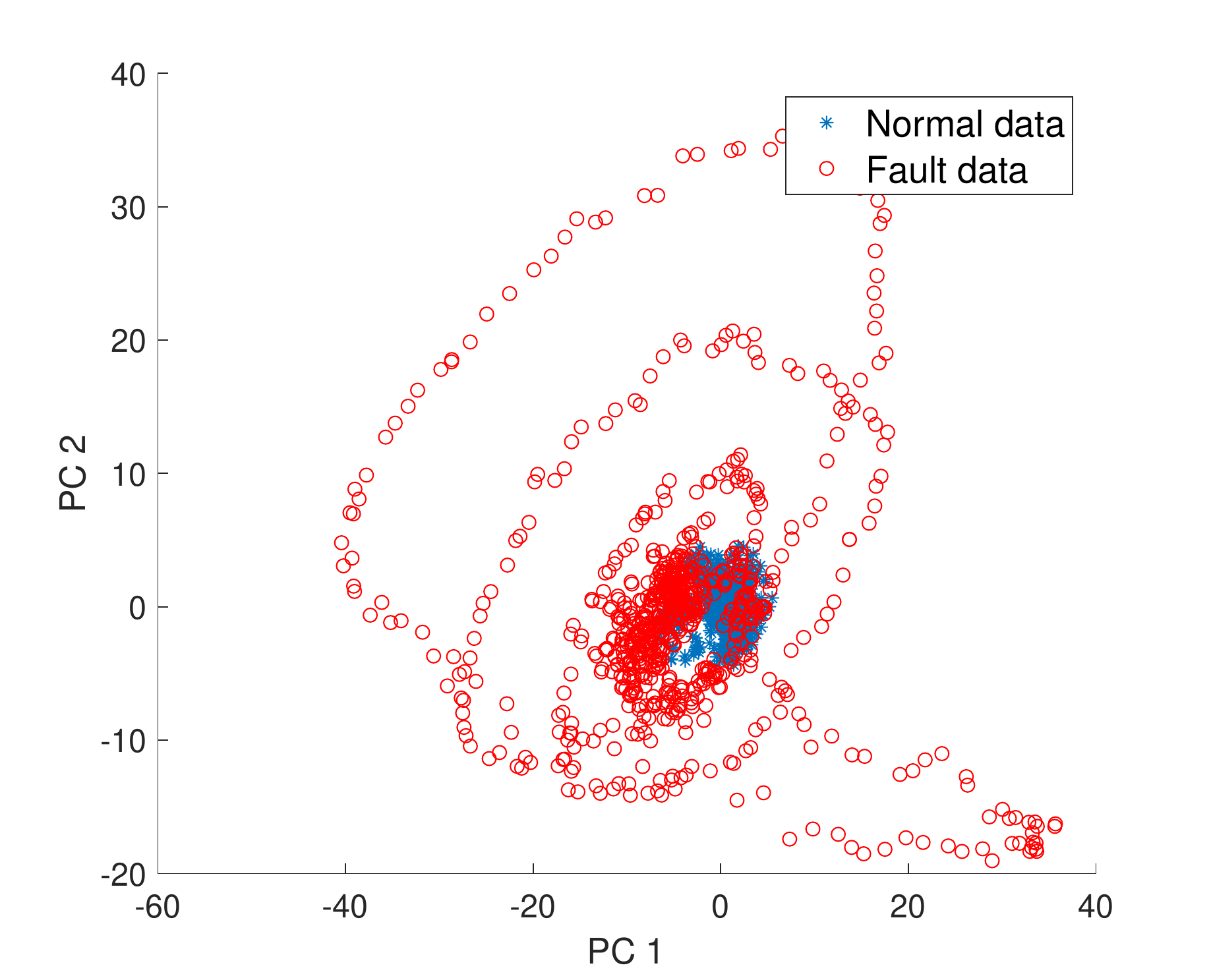}}\subfloat[Visualization plot of KPCA.\label{fig:sub_KPCA_2d}]{\includegraphics[scale=0.25]{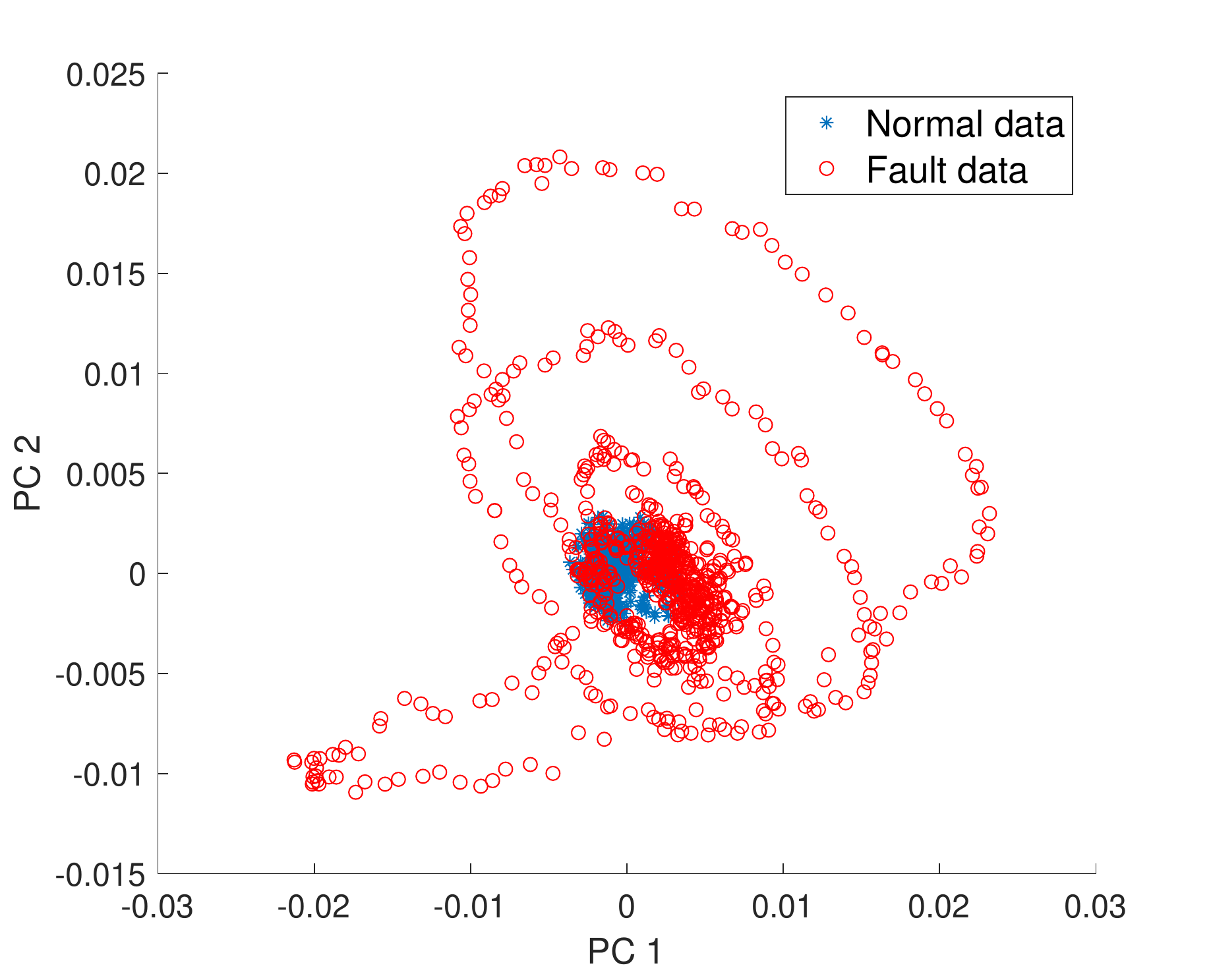}}
\par\end{centering}
\begin{centering}
\subfloat[Visualization plot of autoencoder.\label{fig:sub_ae_2d}]{\includegraphics[scale=0.25]{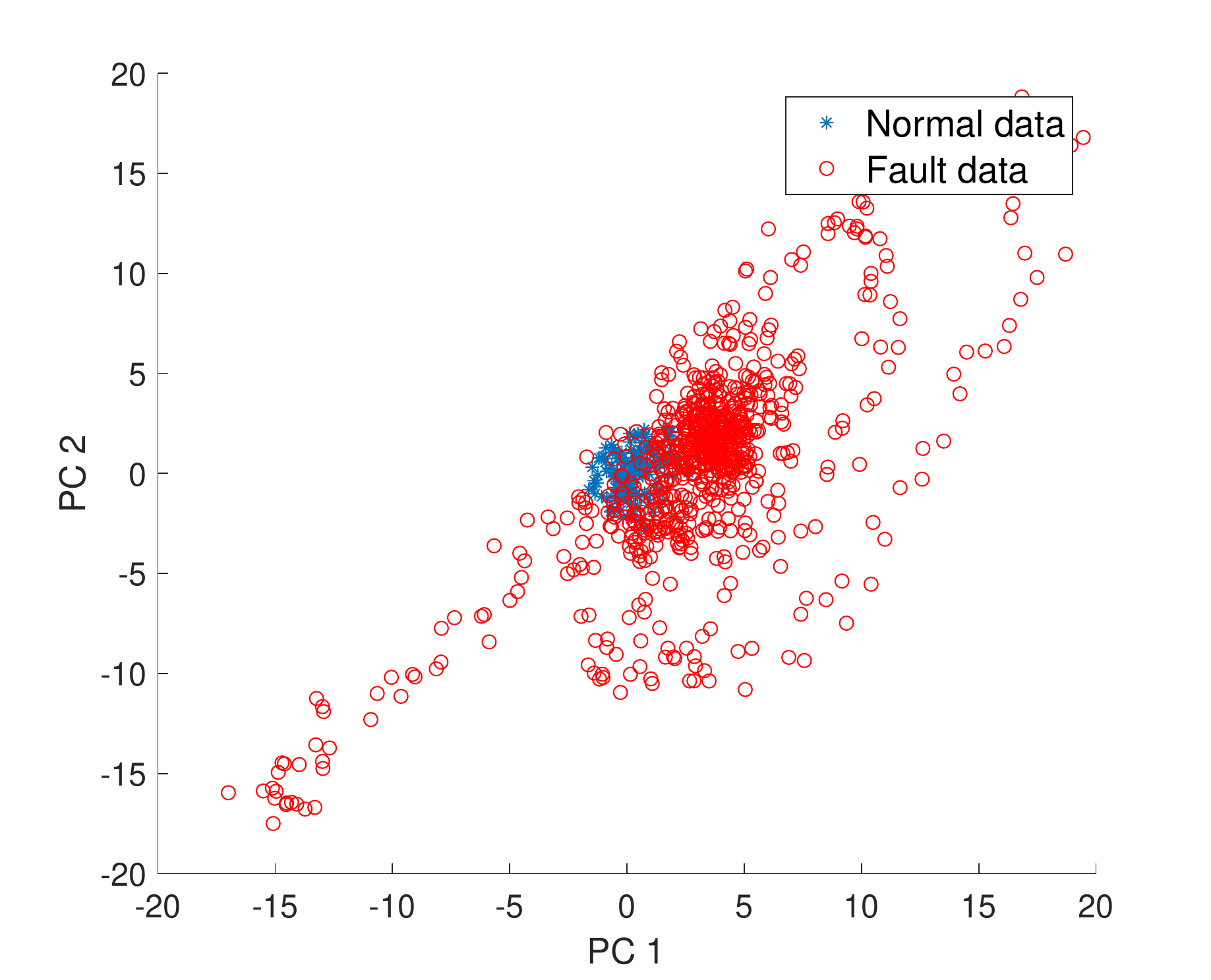}}\subfloat[Visualization plot of NCA.\label{sub_mlp_2d}]{\includegraphics[scale=0.25]{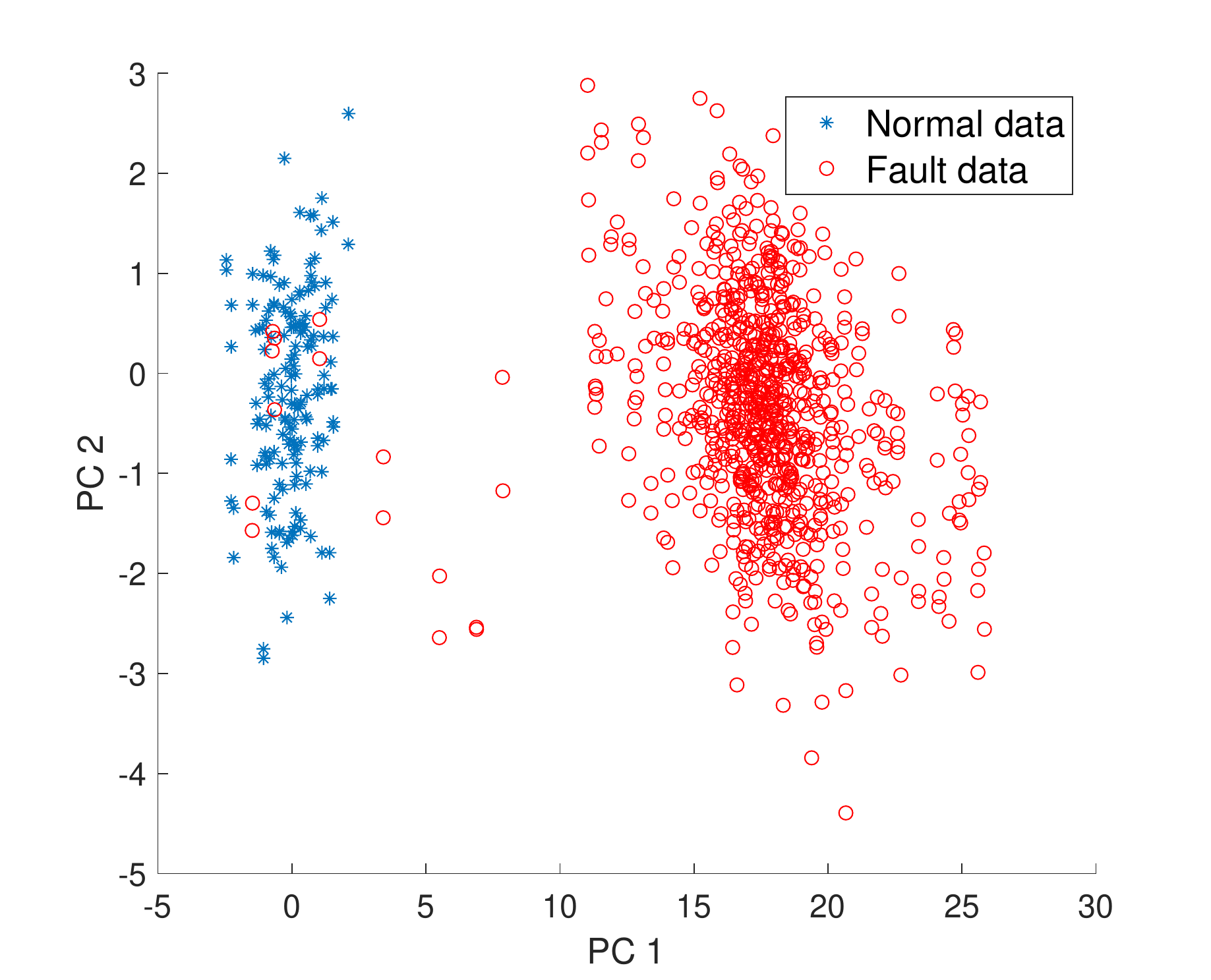}}
\par\end{centering}
\caption{Visualization of the normal samples and samples of Fault 1 on the
first two dimensions of 4 different methods. \label{fig:visofdata}}
\end{figure}

\subsection{Visualization of data of different methods}

In order to intuitively visualize the features of different methods,
we extract 2 features for each method and plot them in Figure \ref{fig:visofdata}.
In Figure \ref{fig:visofdata}, the blue ``$*$'' indicates normal
data, while the red ``$\circ$'' indicates fault data from Fault
1. It can be found that normal data and fault data are largely overlapped
in Figure \ref{fig:sub_PCA_2d}, \ref{fig:sub_KPCA_2d}, and \ref{fig:sub_ae_2d}.
In this case, PCA, KPCA and autoencoder cannot find the significant
information for fault detection. However, in Figure \ref{sub_mlp_2d},
the overlapping of normal samples and fault samples is much less than
those in Figure \ref{fig:sub_PCA_2d}, \ref{fig:sub_KPCA_2d}, and
\ref{fig:sub_ae_2d}. Through the nonlinear feature extraction based
on orthogonal constraints, NCA directly gives out the key features
in a two-dimensional space which include significant information for
fault detection.

\begin{figure}
\subfloat[Illustration of $B^{T}B$.\label{fig:sub_inner_1}]{\includegraphics[scale=0.25]{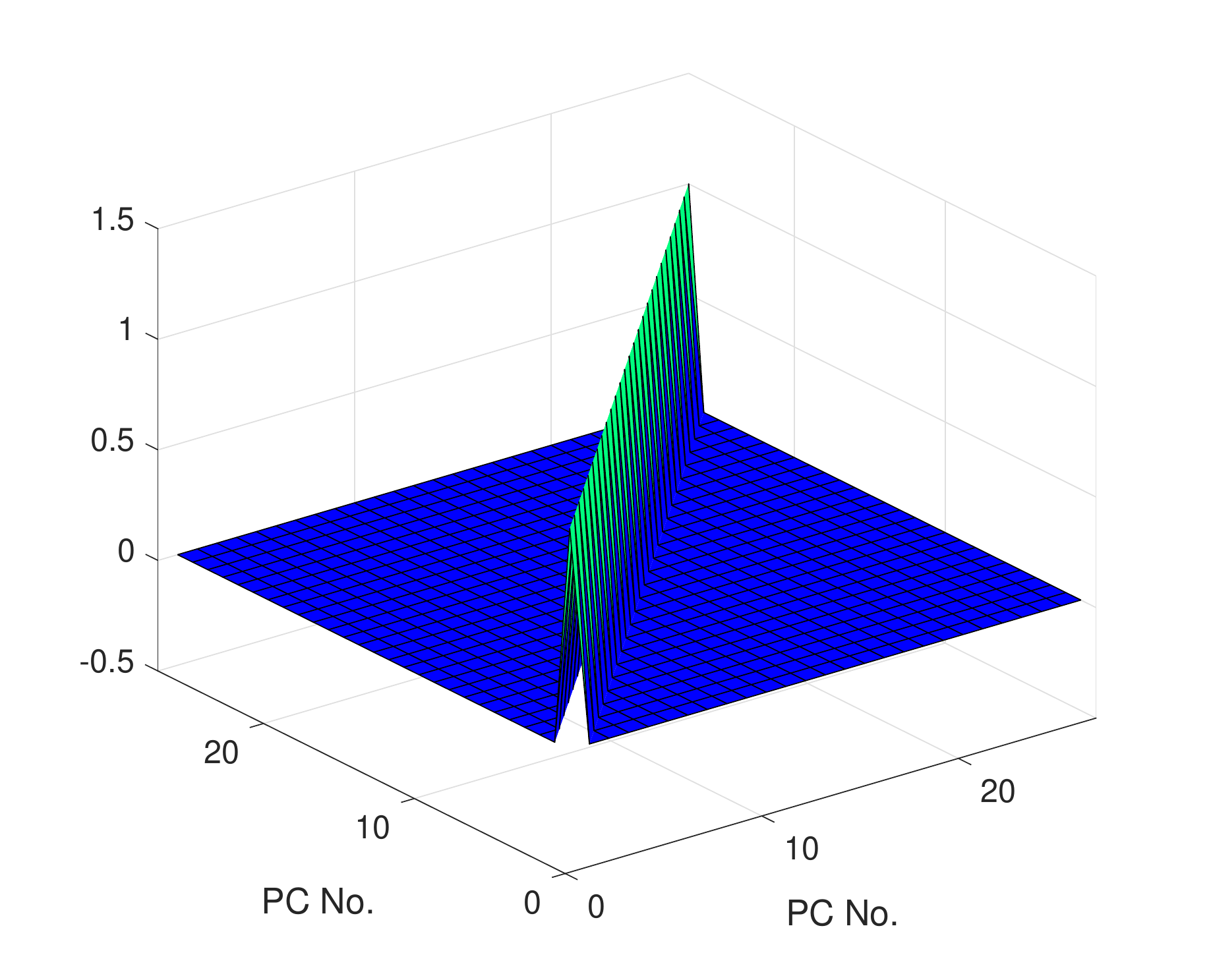}}\subfloat[Illustration of $B^{T}U$.\label{fig:sub_inner_2}]{\includegraphics[scale=0.25]{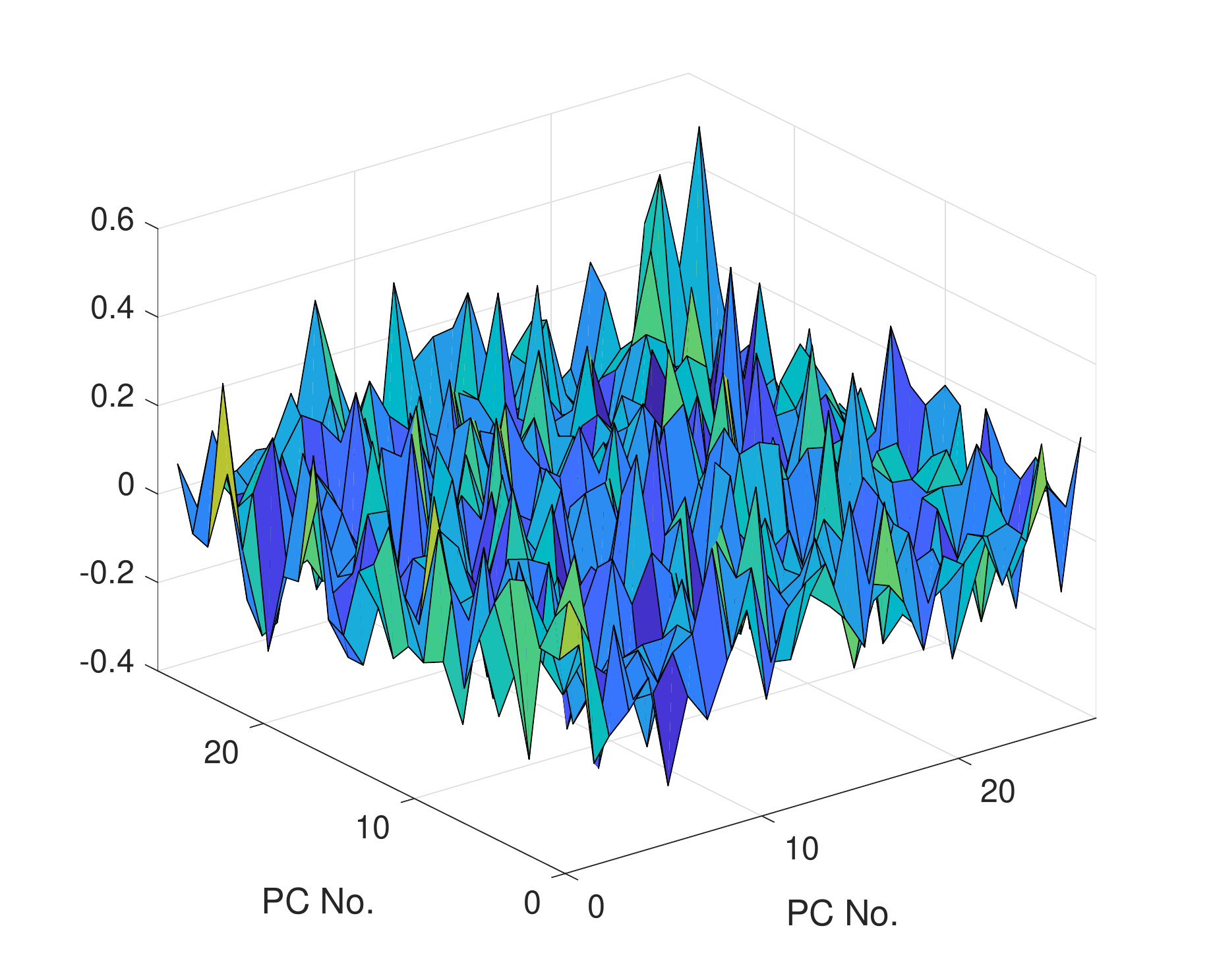}}\caption{Illustration of the correlations of the columns of $B$ and the correlations
between the columns of $B$ and those of $U$.\label{fig:inner_product}}
\end{figure}

\selectlanguage{english}%
Figure \ref{fig:inner_product} shows the difference between $B$
of NCA and the loading matrix $U$ of PCA. Figure \ref{fig:sub_inner_1}
shows the result of $B^{T}B$. It is easy to find that $B$ contains
orthogonormal columns, i.e. ${\bf \mathfrak{b}}_{i}^{T}{\bf \mathfrak{b}}_{j}=1$
if $i=j$, otherwise, ${\bf \mathfrak{b}}_{i}^{T}{\bf \mathfrak{b}}_{j}=0$.
Moreover, the correlations between the columns of $B$ and those of
$P$ are plotted in Figure \ref{fig:sub_inner_2}. Obviously with
the nonlinear feature extract, the optimum solution $B$ of NCA is
utterly different from the loading matrix $U$ of PCA.

\begin{figure}
\centering{}\includegraphics[scale=0.4]{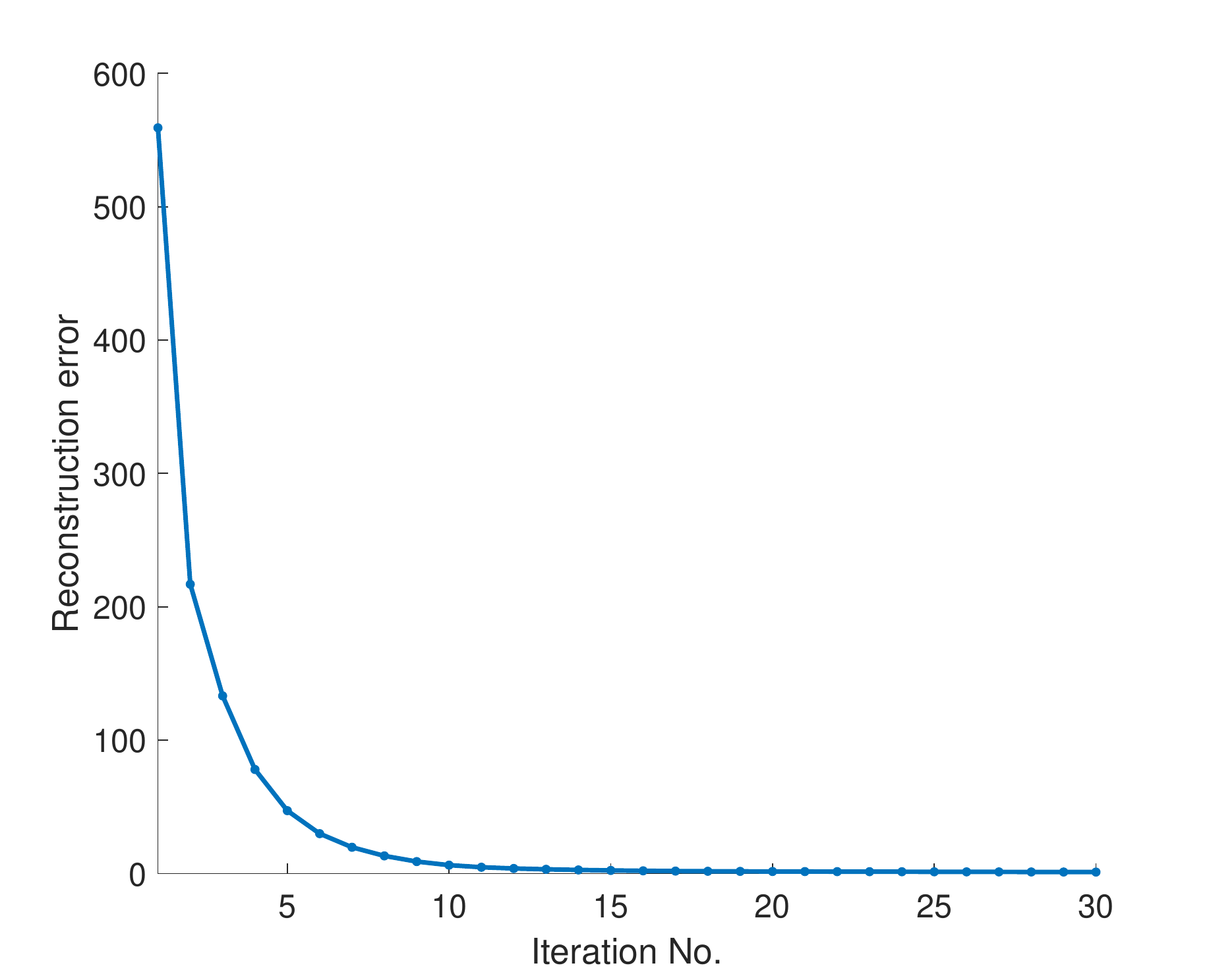}\caption{Convergence plot of NCA.\label{fig:convergence}}
\end{figure}

Figure \ref{fig:convergence} plots the reconstruction errors $\left\Vert X-GB^{T}\right\Vert _{F}^{2}$
in the training stage. It can be found that NCA converges very fast.
Thanks for GPU-accelerated computing, the total training time of NCA
is 2.58 seconds. We perform the experiment on a computer with Intel
Core i7 3.4GHz, 16G RAM, and NVIDIA GeForce GTX 1080Ti. For comparison,
KPCA is performed on the same computer and the training time is 1.53
seconds.

\subsection{\textcolor{black}{Case studies}}

\textcolor{black}{In this subsection, we investigate t}he performance
of our proposed NCA. The performance is compared with PCA, KPCA, and
autoencoder. The source codes of NCA and other methods can be found
in \href{https://github.com/haitaozhao/Neural-Component-Analysis.git}{https://github.com/haitaozhao/Neural-Component-Analysis.git}. 

\textcolor{black}{According to the cumulative percentage variance
(CPV) rule, the reduced dimensionality, $p$, was determined as 27
for PCA such that 85\% of the energy in the eigenspetrum (computed
as the sum of eigenvalues) was retained. In order to give a fair comparison,
the same value $p=27$ was used for KPCA, autoencoder and NCA. }

\textcolor{black}{}
\begin{table*}
\textcolor{black}{\caption{Missed Detection Rate (\%) and False Alarm Rate (\%) (shown in parentheses)
of PCA, LPP, LPP\_Markov and DGE in TEP\label{tab:Missed-Detection-Rate}}
}
\centering{}{\footnotesize{}}%
\begin{tabular}{|c|c|c|c|c|c|c|c|c|}
\hline 
\multirow{2}{*}{{\footnotesize{}No.}} & \multicolumn{2}{c|}{{\footnotesize{}PCA}} & \multicolumn{2}{c|}{{\footnotesize{}KPCA}} & \multicolumn{2}{c|}{{\footnotesize{}autoencoder}} & \multicolumn{2}{c|}{{\footnotesize{}NCA}}\tabularnewline
\cline{2-9} 
 & {\footnotesize{}$T^{2}$} & {\footnotesize{}SPE} & {\footnotesize{}$T^{2}$} & {\footnotesize{}SPE} & {\footnotesize{}$T^{2}$} & {\footnotesize{}SPE} & {\footnotesize{}$T^{2}$} & {\footnotesize{}SPE}\tabularnewline
\hline 
\hline 
{\footnotesize{}1} & \textbf{\footnotesize{}0.50}{\footnotesize{}(4.38)} & {\footnotesize{}0.13(20.6)} & {\footnotesize{}0.5(7.50)} & {\footnotesize{}0.63(5.63)} & \textbf{\footnotesize{}0.50}{\footnotesize{}(3.13)} & {\footnotesize{}0.75(0.00)} & \textbf{\footnotesize{}0.50}{\footnotesize{}(0.00)} & {\footnotesize{}0.75(0.00)}\tabularnewline
\hline 
{\footnotesize{}2} & {\footnotesize{}1.63(2.50)} & {\footnotesize{}0.75(18.1)} & {\footnotesize{}1.63(5.00)} & {\footnotesize{}1.75(5.00)} & \textbf{\footnotesize{}1.50}{\footnotesize{}(1.88)} & {\footnotesize{}2.00(0.00)} & {\footnotesize{}1.75(0.00)} & \textbf{\footnotesize{}1.50}{\footnotesize{}(0.00)}\tabularnewline
\hline 
{\footnotesize{}3} & {\footnotesize{}92.0(0.63)} & {\footnotesize{}71.9(30.0)} & {\footnotesize{}88.5(1.25)} & {\footnotesize{}93.8(0.63)} & {\footnotesize{}96.4(1.88)} & {\footnotesize{}99.9(0.00)} & {\footnotesize{}98.4(0.00)} & {\footnotesize{}97.9(0.63)}\tabularnewline
\hline 
{\footnotesize{}4} & {\footnotesize{}39.1(2.50)} & {\footnotesize{}0.00(24.4)} & {\footnotesize{}51.1(5.00)} & {\footnotesize{}81.8(3.75)} & {\footnotesize{}55.8(0.63)} & {\footnotesize{}96.3(0.00)} & {\footnotesize{}63.3(1.25)} & \textbf{\footnotesize{}5.75}{\footnotesize{}(0.63)}\tabularnewline
\hline 
{\footnotesize{}5} & {\footnotesize{}73.8(0.63)} & {\footnotesize{}50.3(22.5)} & {\footnotesize{}70.5(5.00)} & {\footnotesize{}79.3(3.75)} & {\footnotesize{}73.3(0.63)} & {\footnotesize{}77.9(0.00)} & {\footnotesize{}75.4(1.25)} & {\footnotesize{}71.1(1.88)}\tabularnewline
\hline 
{\footnotesize{}6} & {\footnotesize{}1.00(0.00)} & {\footnotesize{}0.00(10.6)} & {\footnotesize{}0.88(2.50)} & {\footnotesize{}2.88(3.13)} & {\footnotesize{}1.00(0.00)} & {\footnotesize{}1.13(0.00)} & \textbf{\footnotesize{}0.50}{\footnotesize{}(0.00)} & {\footnotesize{}0.88(0.00)}\tabularnewline
\hline 
{\footnotesize{}7} & \textbf{\footnotesize{}0.00}{\footnotesize{}(0.00)} & {\footnotesize{}0.00(16.3)} & \textbf{\footnotesize{}0.00}{\footnotesize{}(0.63)} & {\footnotesize{}2.0(0.63)} & \textbf{\footnotesize{}0.00}{\footnotesize{}(0.00)} & {\footnotesize{}30.1(0.63)} & \textbf{\footnotesize{}0.00}{\footnotesize{}(0.00)} & \textbf{\footnotesize{}0.00}{\footnotesize{}(1.88)}\tabularnewline
\hline 
{\footnotesize{}8} & \textbf{\footnotesize{}2.50}{\footnotesize{}(0.63)} & {\footnotesize{}1.75(17.5)} & \textbf{\footnotesize{}2.50}{\footnotesize{}(2.50)} & {\footnotesize{}4.75(3.13)} & \textbf{\footnotesize{}2.50}{\footnotesize{}(1.88)} & {\footnotesize{}4.38(0.00)} & {\footnotesize{}2.75(0.00)} & \textbf{\footnotesize{}2.50}{\footnotesize{}(2.50)}\tabularnewline
\hline 
{\footnotesize{}9} & {\footnotesize{}96.4(5.00)} & {\footnotesize{}76.9(23.8)} & {\footnotesize{}89.0(15.0)} & {\footnotesize{}94.1(5.63)} & {\footnotesize{}96.0(5.00)} & {\footnotesize{}99.38(1.88)} & {\footnotesize{}97.9(2.50)} & {\footnotesize{}94.0(5.63)}\tabularnewline
\hline 
{\footnotesize{}10} & {\footnotesize{}58.4(0.00)} & {\footnotesize{}24.13(15)} & \textbf{\footnotesize{}48.9}{\footnotesize{}(1.88)} & {\footnotesize{}81.6(0.63)} & {\footnotesize{}64.3(0.63)} & {\footnotesize{}77.38(0.00)} & {\footnotesize{}66.3(0.00)} & {\footnotesize{}72.8(1.88)}\tabularnewline
\hline 
{\footnotesize{}11} & {\footnotesize{}47.9(0.63)} & {\footnotesize{}19.0(20.0)} & \textbf{\footnotesize{}38.6}{\footnotesize{}(4.38)} & {\footnotesize{}52.4(1.88)} & {\footnotesize{}49.6(1.25)} & {\footnotesize{}71.9(0.00)} & {\footnotesize{}52.0(0.63)} & {\footnotesize{}30.1(8.13)}\tabularnewline
\hline 
{\footnotesize{}12} & {\footnotesize{}1.25(0.63)} & {\footnotesize{}1.13(22.5)} & \textbf{\footnotesize{}1.00}{\footnotesize{}(3.13)} & {\footnotesize{}10.4(0.63)} & \textbf{\footnotesize{}1.00}{\footnotesize{}(1.88)} & {\footnotesize{}2.63(0.00)} & \textbf{\footnotesize{}1.00}{\footnotesize{}(1.25)} & {\footnotesize{}5.50(7.5)}\tabularnewline
\hline 
{\footnotesize{}13} & \textbf{\footnotesize{}4.88}{\footnotesize{}(0.00)} & {\footnotesize{}3.75(12.5)} & {\footnotesize{}4.63(1.25)} & {\footnotesize{}18.0(0.63)} & {\footnotesize{}5.75(0.63)} & {\footnotesize{}5.88(1.25)} & {\footnotesize{}5.13(1.25)} & \textbf{\footnotesize{}4.88}{\footnotesize{}(3.75)}\tabularnewline
\hline 
{\footnotesize{}14} & {\footnotesize{}0.13(0.63)} & {\footnotesize{}0.0(23.13)} & \textbf{\footnotesize{}0.00}{\footnotesize{}(1.25)} & {\footnotesize{}9.88(0.63)} & {\footnotesize{}0.38(0.63)} & {\footnotesize{}4.88(0.00)} & {\footnotesize{}0.38(0.63)} & \textbf{\footnotesize{}0.00}{\footnotesize{}(5.00)}\tabularnewline
\hline 
{\footnotesize{}15} & {\footnotesize{}95.1(0.00)} & {\footnotesize{}74.4(16.9)} & {\footnotesize{}86.9(3.13)} & {\footnotesize{}94.1(2.50)} & {\footnotesize{}94.3(0.63)} & {\footnotesize{}98.1(0.00)} & {\footnotesize{}97.5(0.00)} & {\footnotesize{}90.9(3.75)}\tabularnewline
\hline 
{\footnotesize{}16} & {\footnotesize{}76.8(6.88)} & {\footnotesize{}30.8(22.5)} & {\footnotesize{}64.4(20.6)} & {\footnotesize{}86.9(6.25)} & {\footnotesize{}82.1(5.00)} & {\footnotesize{}91.0(1.88)} & {\footnotesize{}85.6(1.88)} & {\footnotesize{}76.6(1.25)}\tabularnewline
\hline 
{\footnotesize{}17} & {\footnotesize{}19.9(1.25)} & {\footnotesize{}2.50(25.6)} & \textbf{\footnotesize{}13.5}{\footnotesize{}(0.63)} & {\footnotesize{}28.0(0.00)} & {\footnotesize{}20.1(1.88)} & {\footnotesize{}27.4(0.00)} & {\footnotesize{}15.4(0.00)} & {\footnotesize{}53.3(0.00)}\tabularnewline
\hline 
{\footnotesize{}18} & {\footnotesize{}10.9(0.0)} & {\footnotesize{}6.25(20.6)} & {\footnotesize{}10.0(2.50)} & {\footnotesize{}14.4(2.50)} & {\footnotesize{}11.0(3.75)} & {\footnotesize{}11.63(0.63)} & \textbf{\footnotesize{}9.88}{\footnotesize{}(0.63)} & {\footnotesize{}12.1(0.00)}\tabularnewline
\hline 
{\footnotesize{}19} & {\footnotesize{}93.25(0.0)} & {\footnotesize{}41.8(14.4)} & {\footnotesize{}82.5(1.88)} & {\footnotesize{}87.3(0.63)} & {\footnotesize{}91.8(0.00)} & {\footnotesize{}99.8(0.00)} & {\footnotesize{}100(0.00)} & {\footnotesize{}99.4(0.00)}\tabularnewline
\hline 
{\footnotesize{}20} & {\footnotesize{}62.63(0.0)} & {\footnotesize{}26.9(15.6)} & {\footnotesize{}52.1(2.50)} & {\footnotesize{}90.0(2.50)} & {\footnotesize{}64.1(0.00)} & {\footnotesize{}78.1(0.00)} & {\footnotesize{}69.1(0.00)} & {\footnotesize{}52.0(0.00)}\tabularnewline
\hline 
{\footnotesize{}21} & {\footnotesize{}62.3(1.25)} & {\footnotesize{}33.5(31.3)} & {\footnotesize{}59.5(5.63)} & {\footnotesize{}71.0(3.75)} & {\footnotesize{}61.6(1.25)} & {\footnotesize{}79.6(0.63)} & {\footnotesize{}64.8(0.00)} & {\footnotesize{}78.0(0.00)}\tabularnewline
\hline 
\end{tabular}{\footnotesize \par}
\end{table*}

\textcolor{black}{Missed detection rate (MDR) refers to the rate of
the abnormal events being falsely identified as normal events in the
monitoring process, which is only applied to the fault detection situation.
For testing all the 21 faults, MDR is recorded together in Table \ref{tab:Missed-Detection-Rate},
where smaller values indicate better performances. False alarm rate
(FAR) which refers to the normal process monitoring results of PCA,
KPCA, autoencoder and NCA are shown in parentheses. The small FARs
also indicate better performances. In Table \ref{tab:Missed-Detection-Rate},
the best achieved performance for each fault is highlighted in bold.
In this study, we only consider the fault cases where MDR$<50\%$
and FAR$\leq5\%$. Because if MDR$\geq50\%$, the detection performance
would be even worse than random guess whose MDR is 50\%. Moreover,
we adopt the threshold value for FAR as 5\% which is commonly used
in fault detection.  }

\textcolor{black}{NCA outperforms PCA, KPCA and autoencoder in 11
cases with lower MDRs. Autoencoder gives the best performance with
5 cases, KPCA gives the best performance with 7 cases and PCA gives
the best performance with 4 cases. Autoencoder, KPCA, and NCA provide
better results than PCA in more fault cases indicates that nonlinear
extensions can generally obtain better fault detection results than
linear PCA. Although both KPCA and NCA include orthogonal constraints
in their feature extraction, the performances of NCA are much better
than KPCA. The reason is that NCA adaptively learns the parameters
of a neural network, while KPCA uses prefixed kernel and the associated
parameter. NCA obtains different parameters for different variables
with nonlinear combination through backpropagation strategy and should
be much more suitable for nonlinear feature extraction and the following
fault detection tasks.}

\begin{figure*}
\begin{centering}
\subfloat[PCA\label{fig:sub_PCA_f2}]{\includegraphics[scale=0.4]{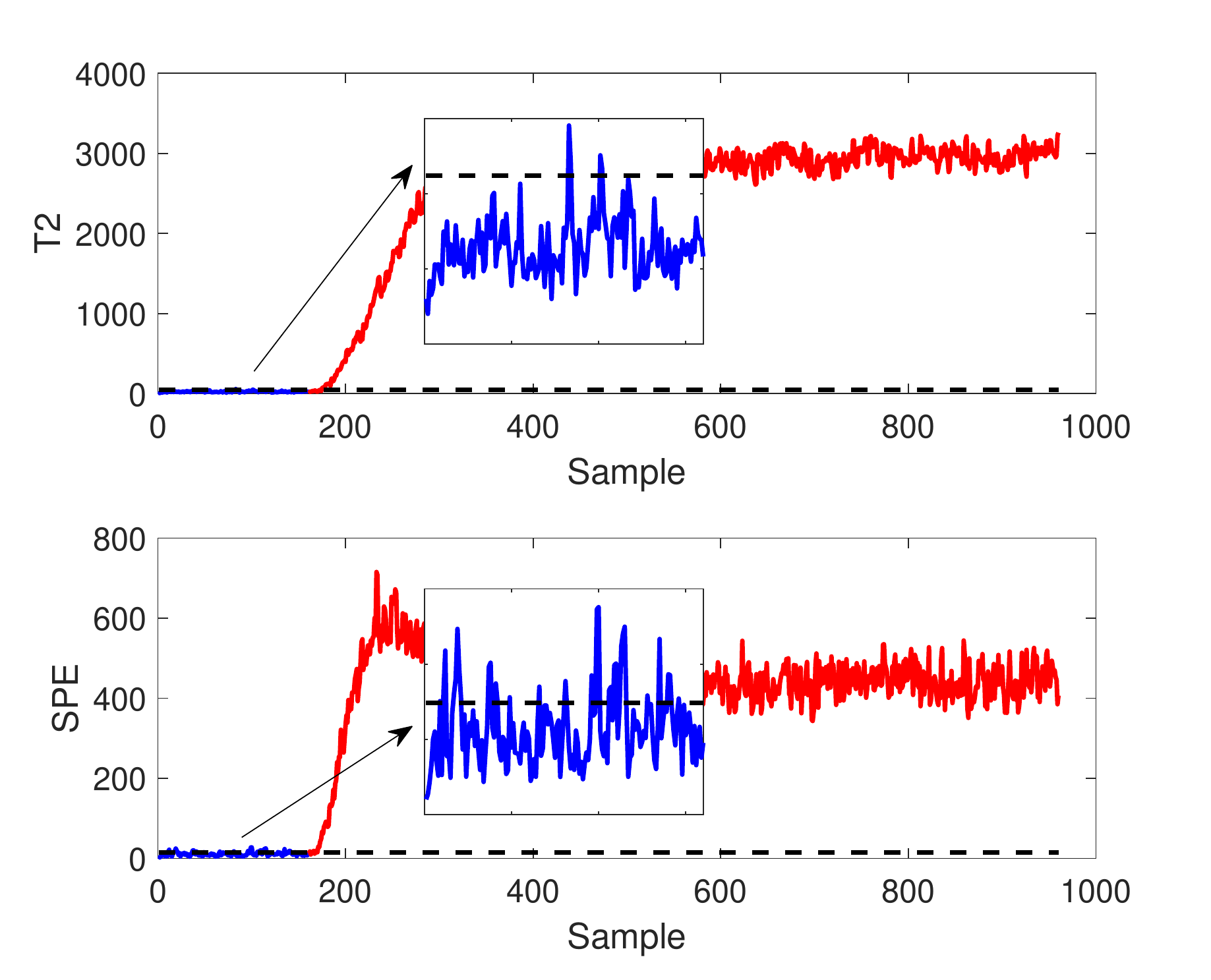}}\subfloat[KPCA\label{fig:sub_KPCA_f2}]{\includegraphics[scale=0.4]{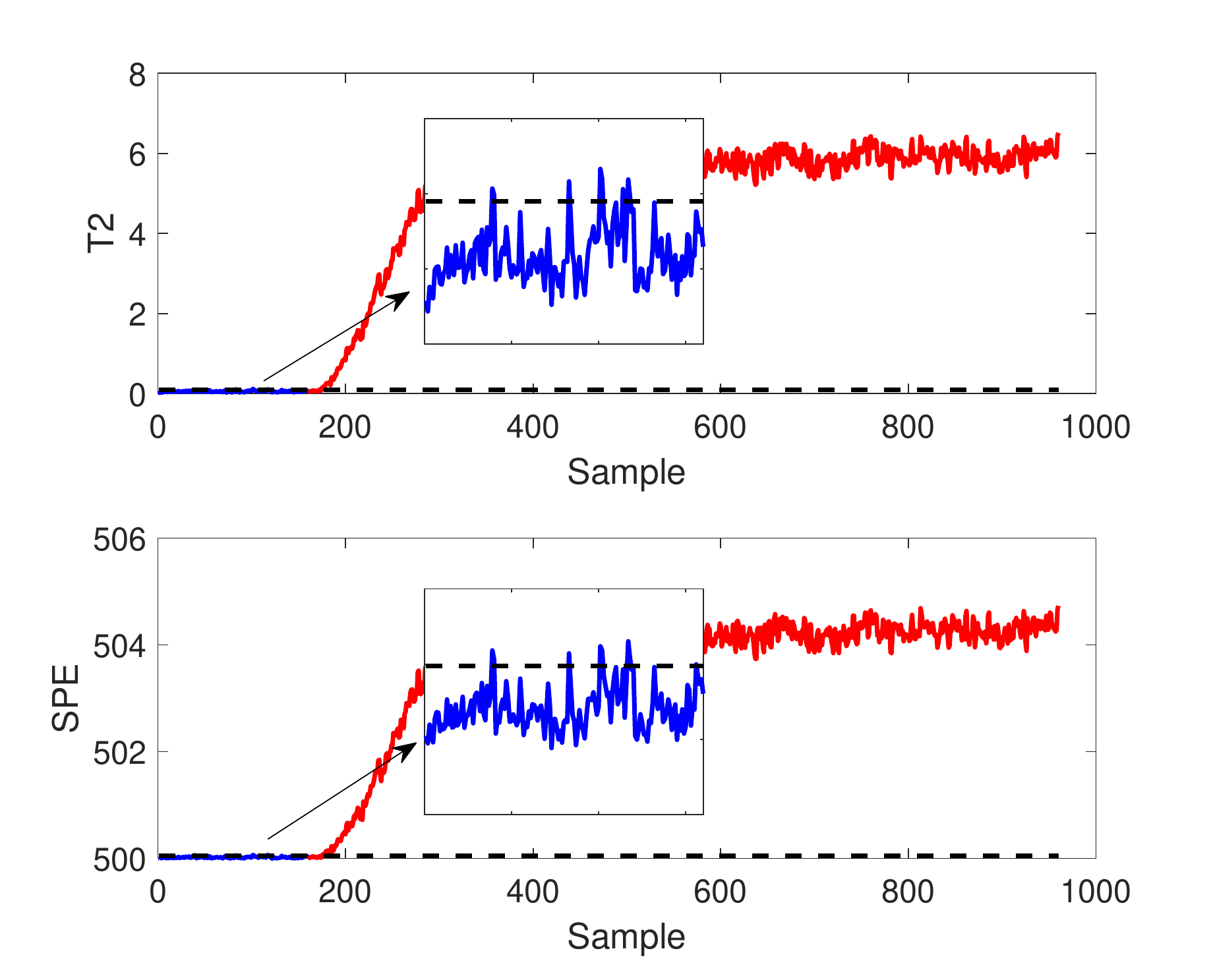}}
\par\end{centering}
\begin{centering}
\subfloat[Autoencoder\label{fig:sub_AE_f2}]{\includegraphics[scale=0.4]{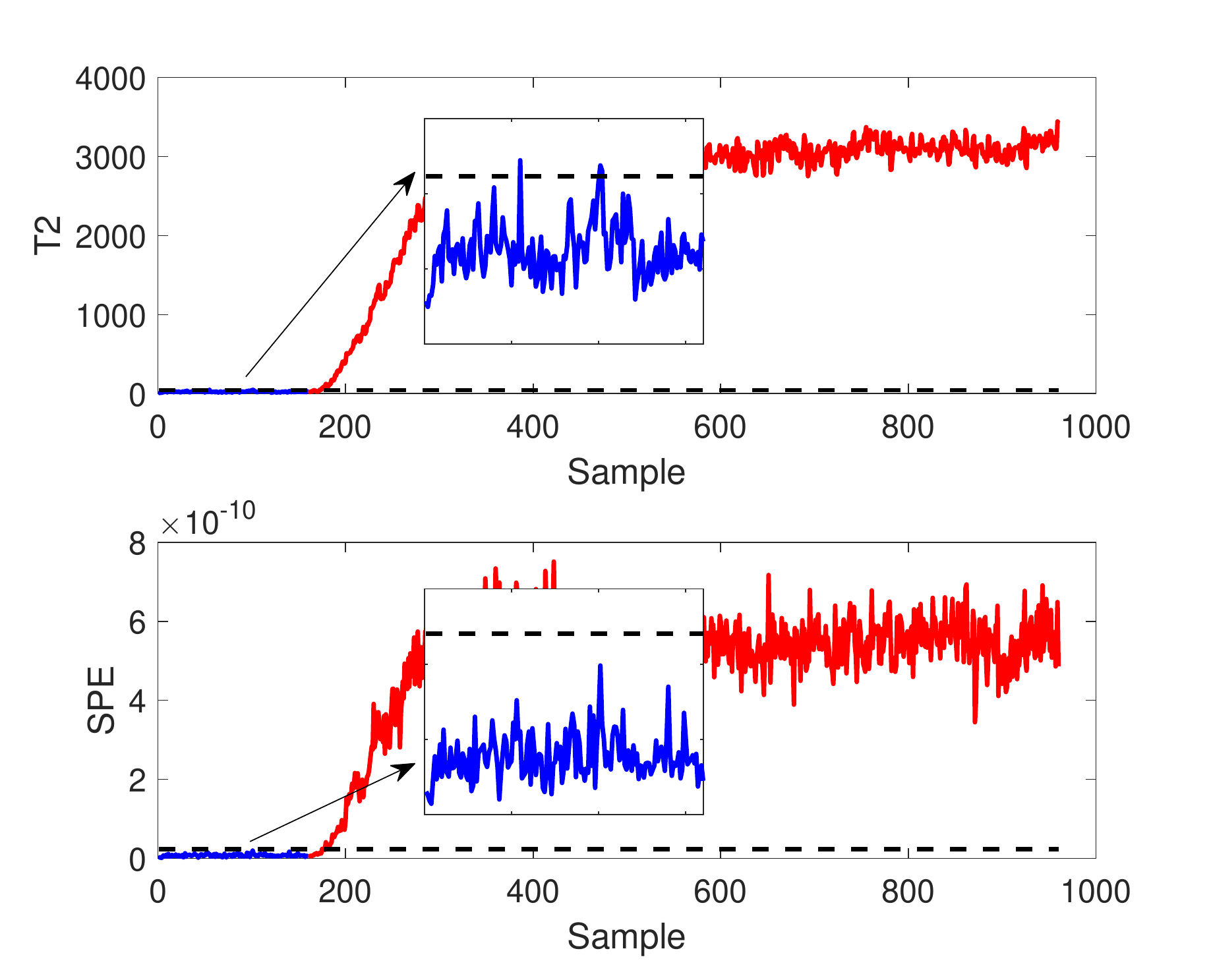}}\subfloat[NCA\label{fig:sub_NCA_f2}]{\includegraphics[scale=0.4]{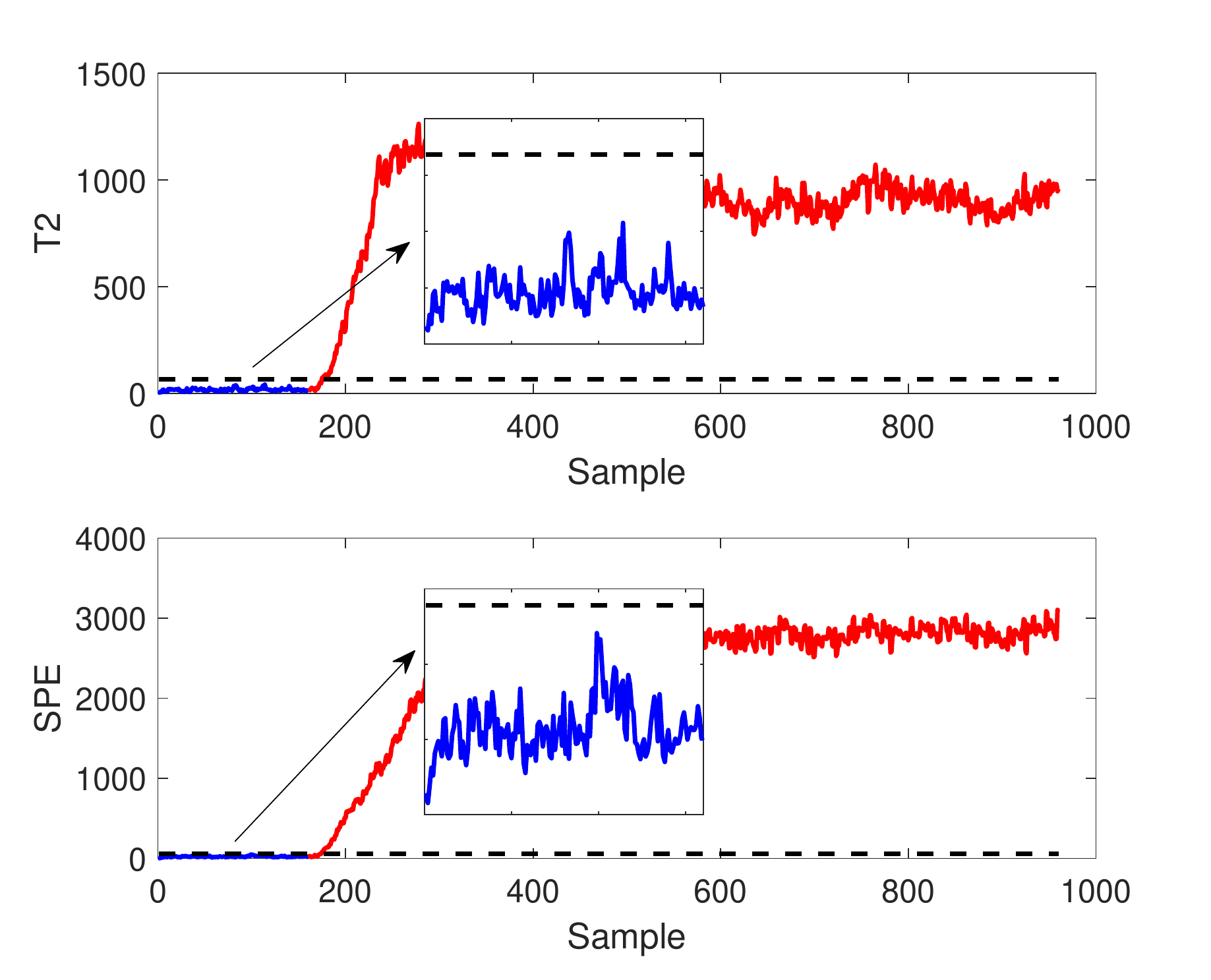}}
\par\end{centering}
\caption{\foreignlanguage{english}{Monitoring results of 4 different methods for \foreignlanguage{british}{Fault
2. \label{fig:fault2_result}}}}
\end{figure*}

\begin{figure*}
\begin{centering}
\subfloat[PCA]{\includegraphics[scale=0.4]{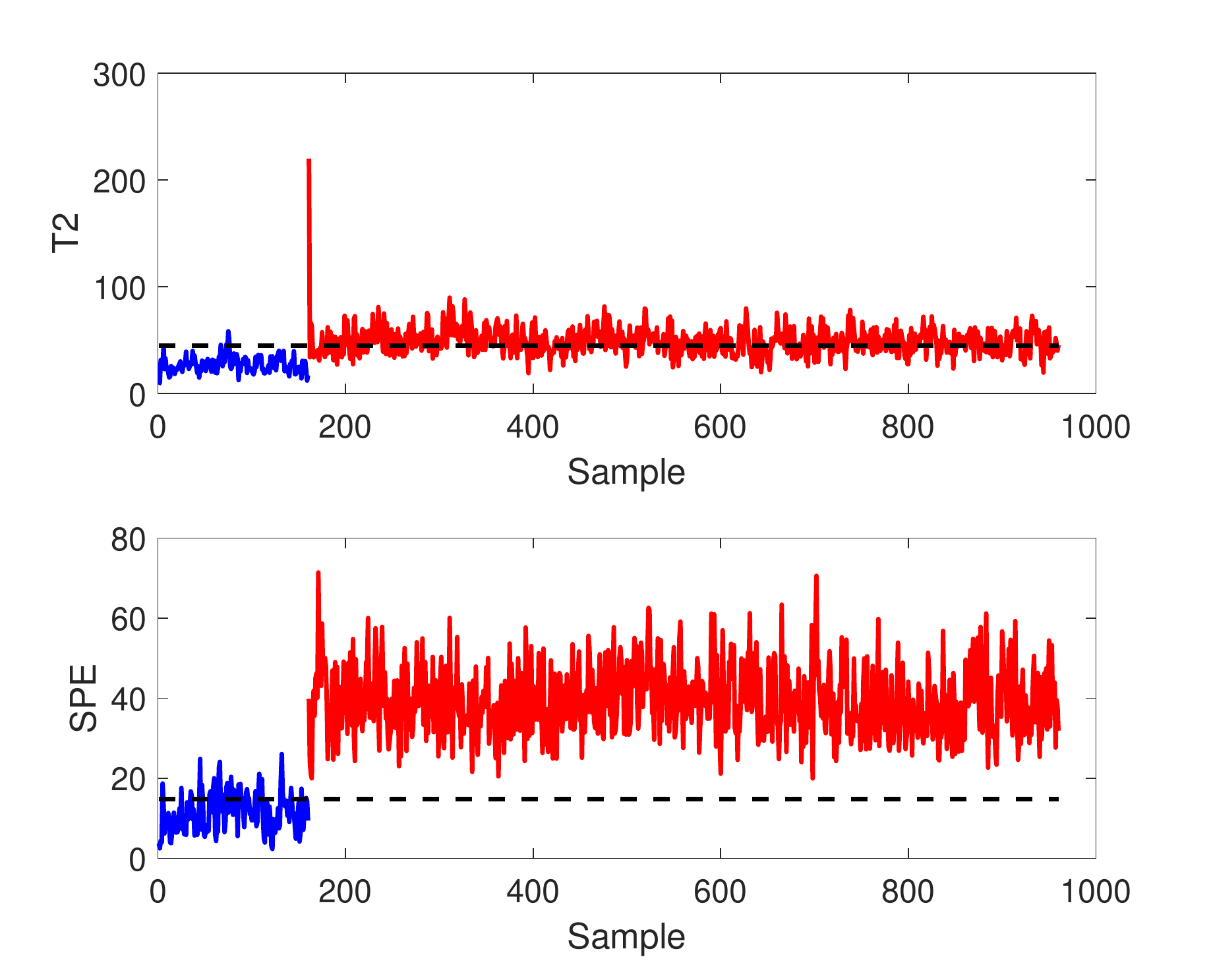}}\subfloat[KPCA]{\includegraphics[scale=0.4]{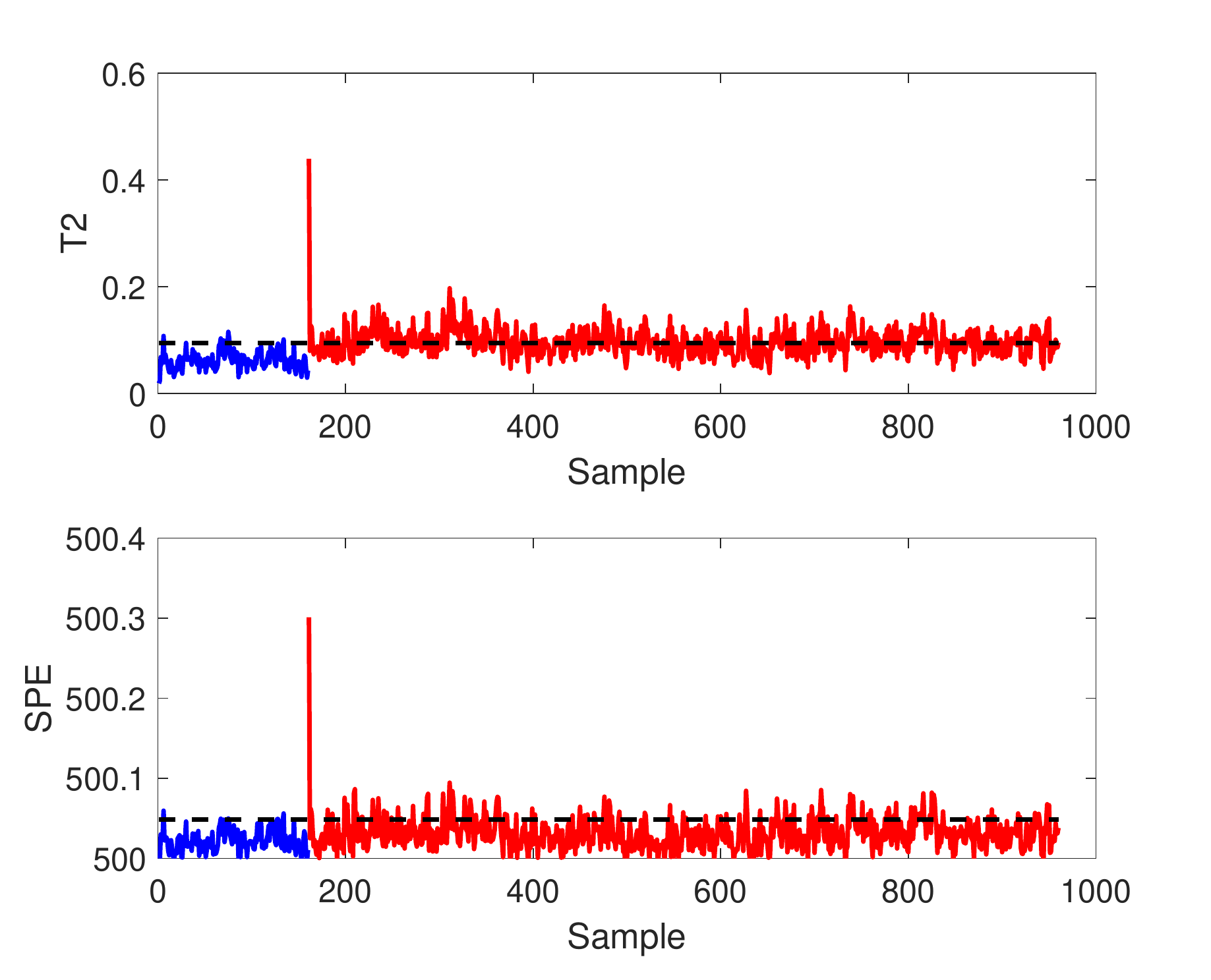}}
\par\end{centering}
\begin{centering}
\subfloat[Autoencoder]{\includegraphics[scale=0.4]{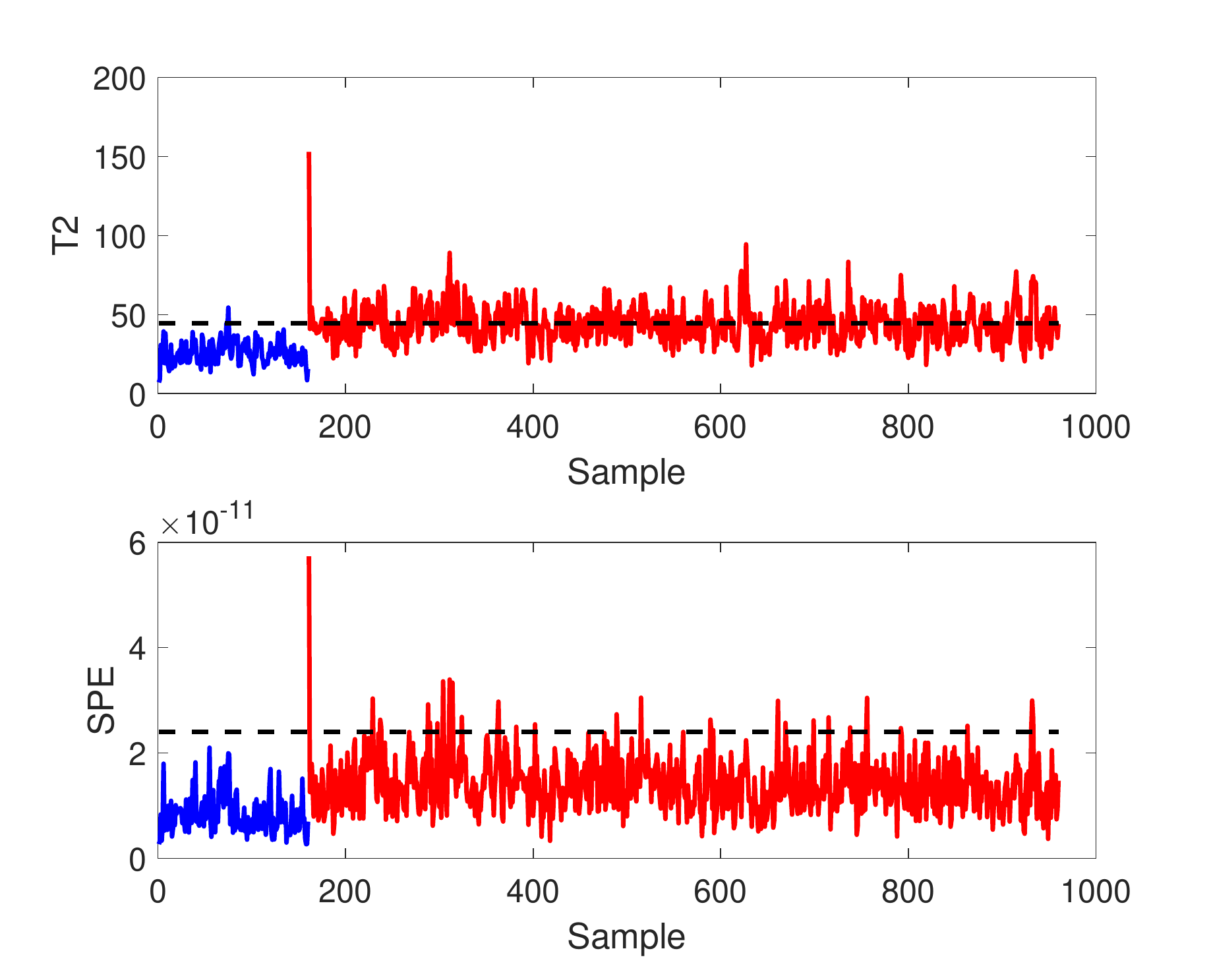}}\subfloat[NCA]{\includegraphics[scale=0.4]{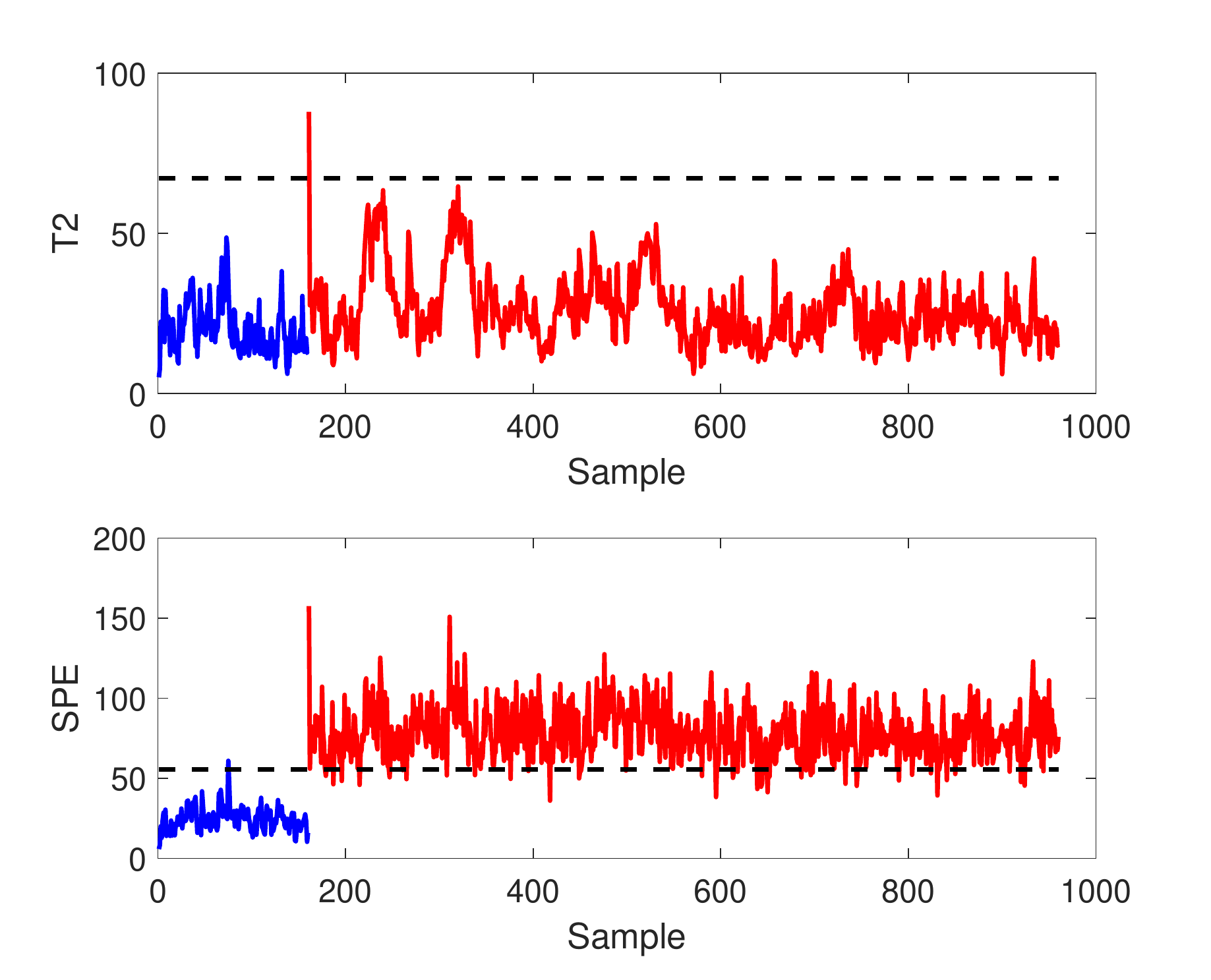}}
\par\end{centering}
\caption{\foreignlanguage{english}{Monitoring results of 4 different methods for \foreignlanguage{british}{Fault
4. \label{fig:fault4_result}}}}
\end{figure*}

\begin{figure*}
\begin{centering}
\subfloat[PCA]{\includegraphics[scale=0.4]{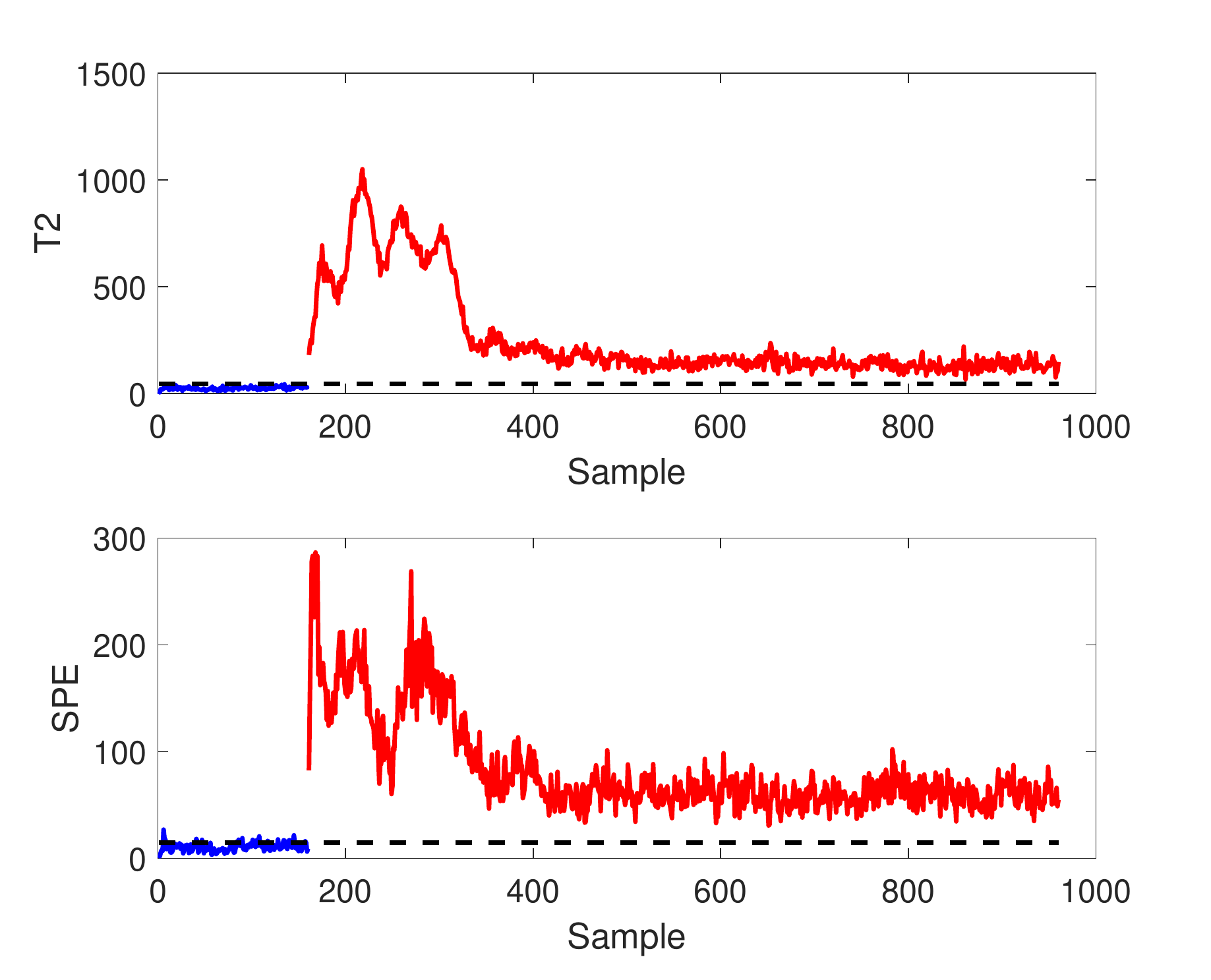}}\subfloat[KPCA]{\includegraphics[scale=0.4]{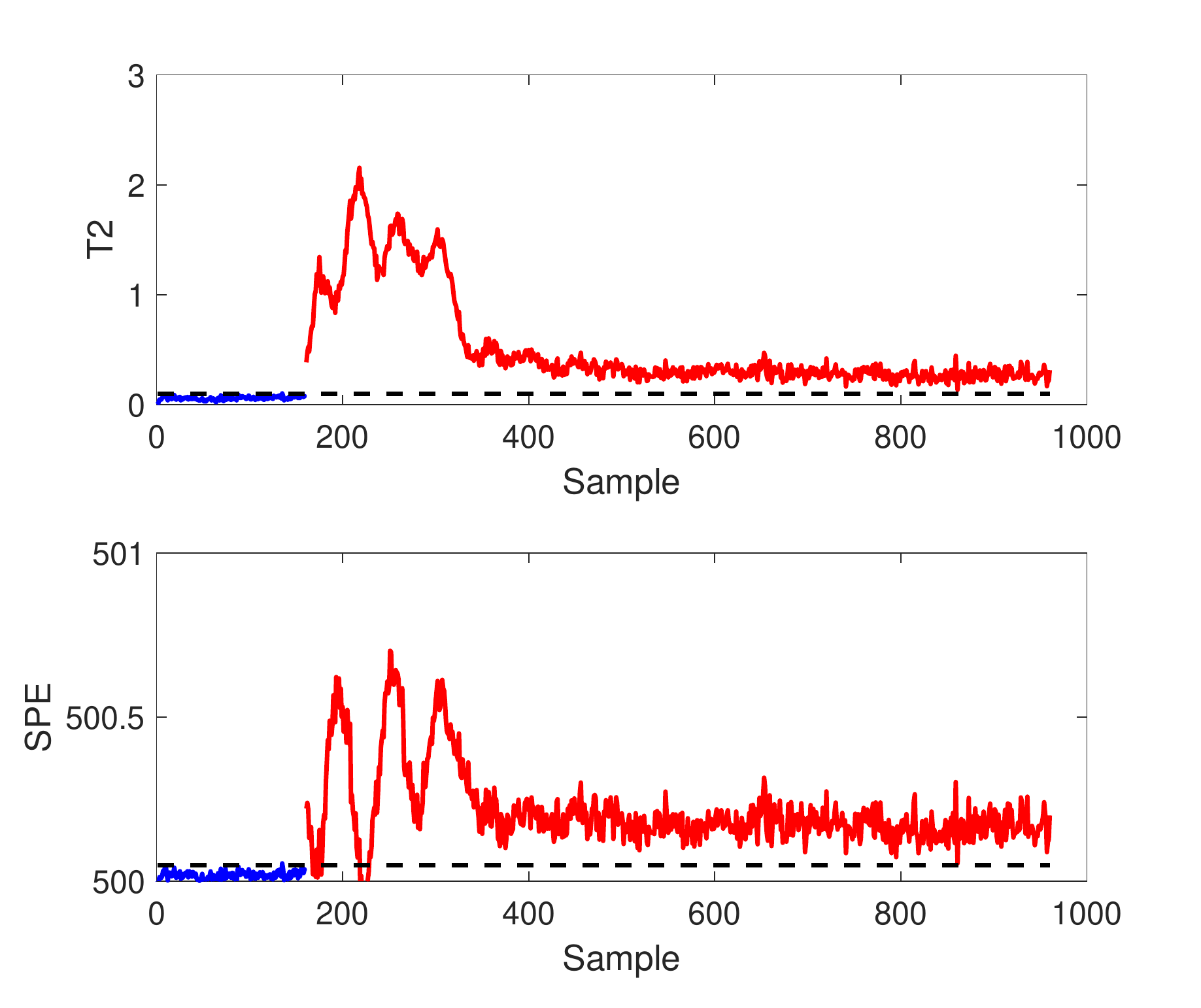}}
\par\end{centering}
\begin{centering}
\subfloat[Autoencoder]{\includegraphics[scale=0.4]{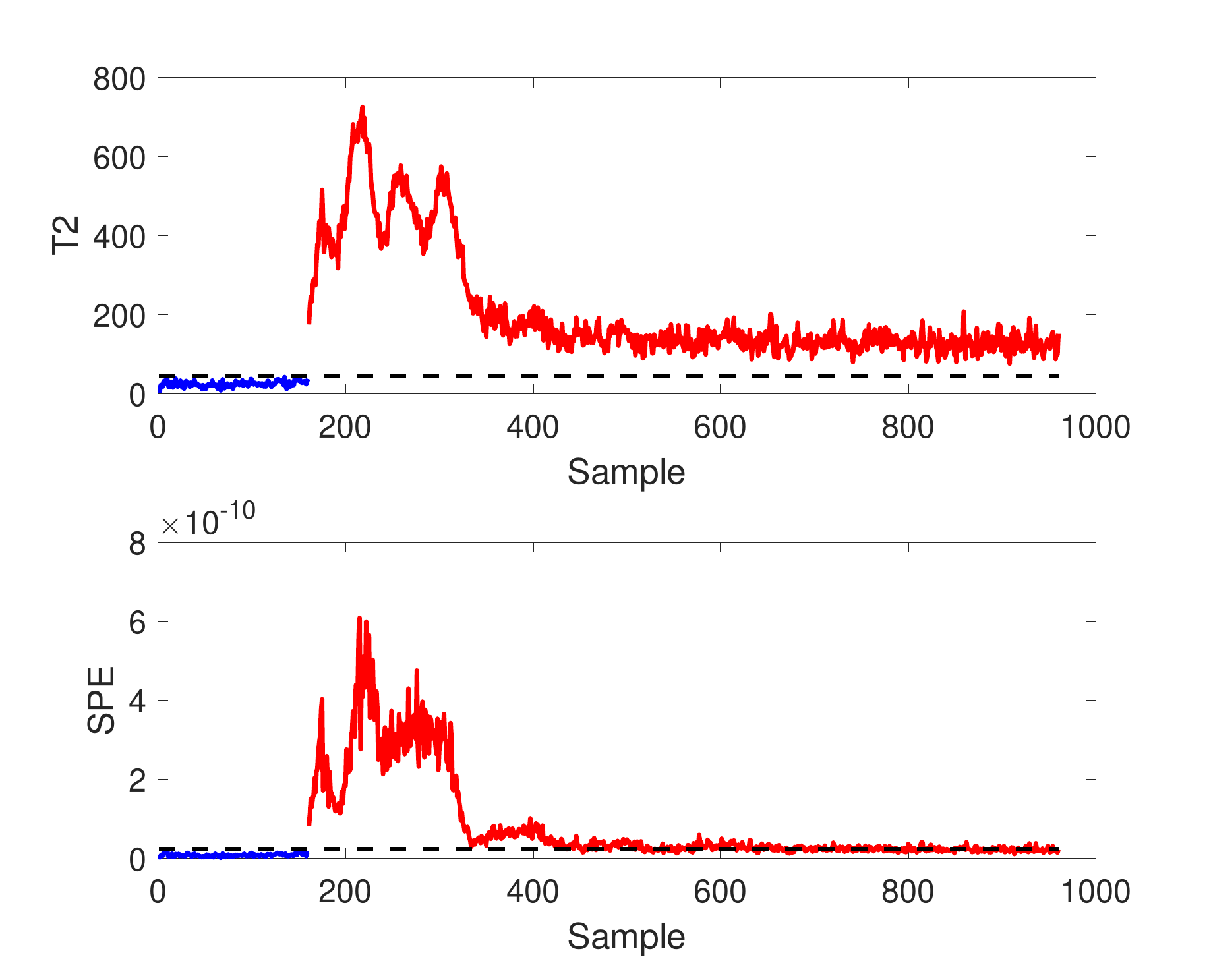}}\subfloat[NCA]{\includegraphics[scale=0.4]{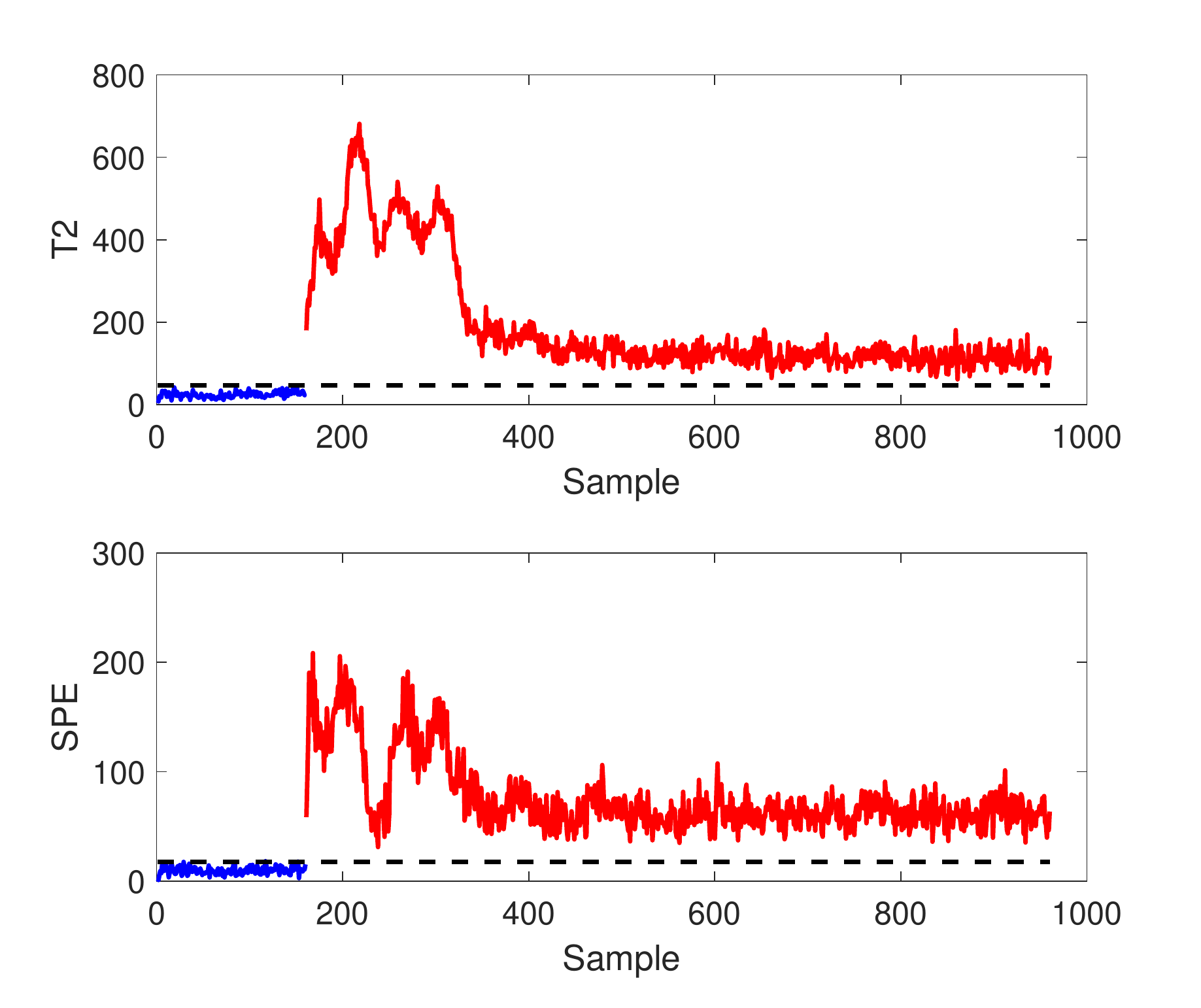}}
\par\end{centering}
\caption{\foreignlanguage{english}{Monitoring results of 4 different methods for \foreignlanguage{british}{Fault
7. \label{fig:fault7_result}}}}
\end{figure*}

\textcolor{black}{Figure \ref{fig:fault2_result}, \ref{fig:fault4_result}
and \ref{fig:fault7_result} illustrate the detailed fault detection
results of Fault 2, 4 and 7. In these figures, the blue points indicate
the first 160 normal samples, while the red points represent the following
800 fault samples. The black dash lines are the control limits according
to the threshold $\tau$. The blue points above control limits lead
to false alarm, while the red points below control limits cause missed
detection.}

\textcolor{black}{According to Table \ref{tab:Missed-Detection-Rate},
all 4 methods can successfully detect Fault 2. For PCA, the FARs are
2.5\% and 18.1\% for $T^{2}$ statistic and SPE statistic respectively.
The FARs of KPCA are 5\% for both $T^{2}$ statistic and SPE statistic.
Autoencoder achieves no false alarm for SPE statistic and its FAR
for $T^{2}$ statistic is 1.88\%. However, as for our proposed NCA,
the FARs are zero for both two statistics. Figure \ref{fig:fault2_result}
shows the plots the results of four methods on Fault 2. For each method,
the subplot above is the results of $T^{2}$ statistic and the subplot
below is the results of SPE statistic. The small overlay plots clearly
show the results of the first 160 normal samples in each subplot.
Based on Table \ref{tab:Missed-Detection-Rate} and Figure \ref{fig:fault2_result},
for Fault 2, we can found that NCA have the lowest MDR with no false
alarm.}

\selectlanguage{british}%

\selectlanguage{english}%
\textcolor{black}{Figure \ref{fig:fault4_result} illustrates the
results of Fault 4. In this experiment, neither KPCA nor autoencoder
can detect this fault. the MDR of PCA is 39.1\% using $T^{2}$ statistic.
It means that PCA, KPCA, and autoencoder are not suitable for the
detection of Fault 4. However, using SPE statistic, NCA obtains the
MDR of 5.75\% and the FAR of 0.63\%. Obviously NCA is much more appropriate
for detecting Fault 4.}

\selectlanguage{british}%

\selectlanguage{english}%
\textcolor{black}{Figure \ref{fig:fault7_result} shows the detection
results of four methods for Fault 7. For SPE statistic, the MDR of
PCA is 0.00\%, but the FAR is 16.3\% which is greater than 5\%. The
reason for the high FAR is that the linear decomposition of PCA cannot
obtain the appropriate residual subspace to determine the right control
limit. KPCA can detect this fault with the MDR of 2.0\% for SPE statistic
and the FAR is 0.63\%. As for autoencoder, the MDR for SPE statistic
is 30.1\% which is much higher than that of KPCA. NCA outperforms
these two nonlinear methods with no missed detection for SPE statistic
and the FAR is 1.88\%. The cause for the high MDR of autoencoder for
SPE statistic should be that autoencoder overfits the training data
and becomes more prone to accept noises and outliers in the normal
data to construct the significant subspace and residual subspace.
Under this condition, the features of the residual subspace may not
contain enough information to detect the fault samples. Thanks for
orthogonal constraints in KPCA and NCA, both of them can successfully
detect the fault data. }

\selectlanguage{british}%

\selectlanguage{english}%
\textcolor{black}{We summarize the case studies below:}
\begin{enumerate}
\item \textcolor{black}{Although no single method gives optimal performance
for all fault cases consisting of diverse numbers of fault conditions.
NCA outperforms PCA, KPCA, and autoencoder with regard to the number
of best performances and emerges as the clear winner.}
\item \textcolor{black}{Due to the incorporation of orthogonal constraints,
NCA is less prone to overfit training data and performs better than
autoencoder. }
\item \textcolor{black}{Since NCA considers both nonlinear feature extraction
and orthogonal constraints for feature extraction, NCA becomes effective
in fault detection. Although NCA is an iterative method and need more
time for training, given the superior performance of NCA, the trade-off
between training time and performance seems justified. }
\end{enumerate}

\section{Conclusion}

\textcolor{black}{In this paper, we propose a nonlinear feature extraction
method, nerual component analysis (NCA), for fault detection. NCA
is a unified model which includes a nonlinear encoder part and a linear
decoder part. Orthogonal constraints are adopted in NCA which can
alleviate the overfitting problem occurred in autoencoder and improve
performances for fault detection. NCA takes the advantages of} the
backpropogation\textcolor{black}{{} technique and the eigenvalue-based
techniques. The convergence of the iteration scheme of NCA is very
fast. The idea behind NCA is general and can potentially be extended
to other detection or diagnosis problems in process monitoring. }

\textcolor{black}{We compare NCA with other linear and nonlinear fault
detection methods, such as PCA, KPCA, and autoencoder. Based on the
case studies, it is clear that NCA outperforms PCA, KPCA, and autoencoder.
NCA can be considered as an alternative to the prevalent data driven
fault detection techniques.}

Future works will be contributed to the design of new regularization
terms to the optimization problem of NCA in Equation (\ref{eq:nca_obj-2}).
Reconstruction-based feature extraction methods, such as PCA, KPCA,
autoencoder, and NCA, are mainly focus on the global Euclidean structure
of process data and overlook the latent local correlation of process
data. In the future, we will try to design new constraints to ensure
the local information is considered in nonlinear feature extraction.

\section*{\textcolor{black}{Acknowledgment}}

\textcolor{black}{This research is sponsored by National Natural Science
Foundation of China (61375007) and Basic Research Programs of Science
and Technology Commission Foundation of Shanghai (15JC1400600).}

\selectlanguage{british}%
\bibliographystyle{ieeetr}
\bibliography{reference}

\begin{thebibliography}{10}

\bibitem{qin2012survey}
S.~J. Qin, ``Survey on data-driven industrial process monitoring and
  diagnosis,'' {\em Annual Reviews in Control}, vol.~36, no.~2, pp.~220--234,
  2012.

\bibitem{macgregor2012monitoring}
J.~MacGregor and A.~Cinar, ``Monitoring, fault diagnosis, fault-tolerant
  control and optimization: Data driven methods,'' {\em Computers \& Chemical
  Engineering}, vol.~47, pp.~111--120, 2012.

\bibitem{yin2012comparison}
S.~Yin, S.~X. Ding, A.~Haghani, H.~Hao, and P.~Zhang, ``A comparison study of
  basic data-driven fault diagnosis and process monitoring methods on the
  benchmark tennessee eastman process,'' {\em Journal of Process Control},
  vol.~22, no.~9, pp.~1567--1581, 2012.

\bibitem{GeSong-5}
Z.~Q. Ge, Z.~H. Song, and F.~Gao, ``Review of recent research on data-based
  process monitoring,'' {\em Industrial and Engineering Chemistry Research},
  vol.~52, no.~10, pp.~3543--3562, 2013.

\bibitem{FeitalKruger-11}
T.~Feital, U.~Kruger, J.~Dutra, J.~C. Dutra, and E.~L. Lima, ``Modeling and
  performance monitoring of multivariate multimodal processes,'' {\em AIChE
  Journal}, vol.~59, no.~5, pp.~1557 -- 1569, 2013.

\bibitem{askarian2016fault}
M.~Askarian, G.~Escudero, M.~Graells, R.~Zarghami, F.~Jalali-Farahani, and
  N.~Mostoufi, ``Fault diagnosis of chemical processes with incomplete
  observations: A comparative study,'' {\em Computers \& chemical engineering},
  vol.~84, pp.~104--116, 2016.

\bibitem{gao2016improved}
X.~Gao and J.~Hou, ``An improved svm integrated gs-pca fault diagnosis approach
  of tennessee eastman process,'' {\em Neurocomputing}, vol.~174, pp.~906--911,
  2016.

\bibitem{7956215}
Y.~Wang, F.~Sun, and B.~Li, ``Multiscale neighborhood normalization-based
  multiple dynamic pca monitoring method for batch processes with frequent
  operations,'' {\em IEEE Transactions on Automation Science and Engineering},
  vol.~PP, no.~99, pp.~1--12, 2017.

\bibitem{7460929}
T.~J. Rato, J.~Blue, J.~Pinaton, and M.~S. Reis, ``Translation-invariant
  multiscale energy-based pca for monitoring batch processes in semiconductor
  manufacturing,'' {\em IEEE Transactions on Automation Science and
  Engineering}, vol.~14, pp.~894--904, April 2017.

\bibitem{luo201611}
L.~Luo, S.~Bao, J.~Mao, and D.~Tang, ``Nonlinear process monitoring based on
  kernel global-local preserving projections,'' {\em Journal of Process
  Control}, vol.~38, no.~Supplement C, pp.~11--21, 2016.

\bibitem{7310889}
N.~Sheng, Q.~Liu, S.~J. Qin, and T.~Chai, ``Comprehensive monitoring of
  nonlinear processes based on concurrent kernel projection to latent
  structures,'' {\em IEEE Transactions on Automation Science and Engineering},
  vol.~13, pp.~1129--1137, April 2016.

\bibitem{kpcaMANSOURI2016334}
M.~Mansouri, M.~Nounou, H.~Nounou, and N.~Karim, ``Kernel pca-based glrt for
  nonlinear fault detection of chemical processes,'' {\em Journal of Loss
  Prevention in the Process Industries}, vol.~40, no.~Supplement C, pp.~334 --
  347, 2016.

\bibitem{robust_self_supervised}
L.~Jiang, Z.~Song, Z.~Ge, and J.~Chen, ``Robust self-supervised model and its
  application for fault detection,'' {\em Industrial \& Engineering Chemistry
  Research}, vol.~56, no.~26, pp.~7503--7515, 2017.

\bibitem{GE2010676}
Z.~Ge, M.~Zhang, and Z.~Song, ``Nonlinear process monitoring based on linear
  subspace and bayesian inference,'' {\em Journal of Process Control}, vol.~20,
  no.~5, pp.~676 -- 688, 2010.

\bibitem{AssociatedNN}
M.~A. Kramer, ``Nonlinear principal component analysis using autoassociative
  neural networks,'' {\em AIChE Journal}, vol.~37, no.~2, pp.~233--243, 1991.

\bibitem{Hinton504}
G.~E. Hinton and R.~R. Salakhutdinov, ``Reducing the dimensionality of data
  with neural networks,'' {\em Science}, vol.~313, no.~5786, pp.~504--507,
  2006.

\bibitem{olpp_caideng}
D.~Cai, X.~He, J.~Han, and H.~J. Zhang, ``Orthogonal laplacianfaces for face
  recognition,'' {\em IEEE Transactions on Image Processing}, vol.~15,
  pp.~3608--3614, Nov 2006.

\bibitem{LIOU201484}
C.-Y. Liou, W.-C. Cheng, J.-W. Liou, and D.-R. Liou, ``Autoencoder for words,''
  {\em Neurocomputing}, vol.~139, no.~Supplement C, pp.~84 -- 96, 2014.

\bibitem{2013arXiv1312}
D.~P. {Kingma} and M.~{Welling}, ``{Auto-Encoding Variational Bayes},'' {\em
  ArXiv e-prints}, Dec. 2013.

\bibitem{Chicco}
D.~Chicco, P.~Sadowski, and P.~Baldi, ``Deep autoencoder neural networks for
  gene ontology annotation predictions,'' in {\em Proceedings of the 5th ACM
  Conference on Bioinformatics, Computational Biology, and Health Informatics},
  BCB '14, (New York, NY, USA), pp.~533--540, ACM, 2014.

\bibitem{BALDI198953}
P.~Baldi and K.~Hornik, ``Neural networks and principal component analysis:
  Learning from examples without local minima,'' {\em Neural Networks}, vol.~2,
  no.~1, pp.~53 -- 58, 1989.

\bibitem{Jolliffe}
I.~Jolliffe, {\em Principal Component Analysis}.
\newblock Springer Verlag, 1986.

\bibitem{Jolliffe-17}
I.~T. Jolliffe, ``A note on the use of principal components in regression,''
  {\em Journal of the Royal Statistical Society}, vol.~31, no.~3, pp.~300 --
  303, 1982.

\bibitem{5231496}
Y.~Li, Y.~Fu, H.~Li, and S.~W. Zhang, ``The improved training algorithm of back
  propagation neural network with self-adaptive learning rate,'' in {\em 2009
  International Conference on Computational Intelligence and Natural
  Computing}, vol.~1, pp.~73--76, June 2009.

\bibitem{procruste}
D.~G. Kendall, ``A survey of the statistical theory of shape,'' {\em
  Statistical Science}, vol.~4, no.~2, pp.~87--99, 1989.

\bibitem{TenBerge1977}
J.~M.~F. Ten~Berge, ``Orthogonal procrustes rotation for two or more
  matrices,'' {\em Psychometrika}, vol.~42, pp.~267--276, Jun 1977.

\bibitem{kerneldensityestimation}
R.~T. Samuel and Y.~Cao, ``Nonlinear process fault detection and identification
  using kernel pca and kernel density estimation,'' {\em Systems Science \&
  Control Engineering}, vol.~4, no.~1, pp.~165--174, 2016.

\bibitem{chiang2001fault}
L.~H. Chiang, R.~D. Braatz, and E.~L. Russell, {\em Fault detection and
  diagnosis in industrial systems}.
\newblock Springer Science \& Business Media, 2001.

\bibitem{Lyman-42}
P.~Lyman and C.~Georgakis, ``Plant-wide control of the tennessee eastman
  problem,'' {\em Computers and Chemical Engineering}, vol.~19, no.~3, pp.~321
  -- 331, 1995.

\bibitem{1658299}
H.~Zhao, P.~C. Yuen, and J.~T. Kwok, ``A novel incremental principal component
  analysis and its application for face recognition,'' {\em IEEE Transactions
  on Systems, Man, and Cybernetics, Part B (Cybernetics)}, vol.~36,
  pp.~873--886, Aug 2006.

\end{thebibliography}
\selectlanguage{english}%

\end{document}